\pdfoutput=1

\documentclass[11pt]{camel-ai}

\usepackage{times}
\usepackage{latexsym}

\usepackage[T1]{fontenc}

\usepackage[utf8]{inputenc}

\usepackage{microtype}

\usepackage{inconsolata}

\usepackage{amsmath}
\usepackage{graphicx} 
\usepackage[english]{babel} 
\usepackage{booktabs}
\usepackage{float}
\usepackage{makecell}
\providecommand{\subfloat}[2][]{\subcaptionbox{#1}{#2}}
\usepackage{algorithm}
\usepackage{algorithmic}
\usepackage{multirow} 
\usepackage{mdframed} 
\usepackage{xcolor}  
\usepackage{wrapfig}
\usepackage{enumitem}

\usepackage{pifont}

\usepackage{amssymb}

\newcommand{\cmark}{\ding{51}}
\newcommand{\xmark}{\ding{55}}

\definecolor{verylightgray}{rgb}{0.95, 0.95, 0.95}

\title{AVA: Attentive VLM Agent for Mastering StarCraft II}

\author[1,2$\dagger$,*]{Weiyu Ma}
\author[1,*]{Yuqian Fu}
\author[3]{Zecheng Zhang}
\author[4]{Bernard Ghanem}
\author[2]{Guohao Li}

\affiliation[1]{Institute of Automation, Chinese Academy of Sciences}
\affiliation[2]{CAMEL-AI.org}
\affiliation[3]{Strukto.ai}
\affiliation[4]{KAUST}

\contribution[*]{Equal contribution.}

\correspondence{\email{sc2meisah@gmail.com}, \email{fuyuqian2022@ia.ac.cn},
\email{guohao.li@eigent.ai}}

\abstract{We introduce \textbf{AVACraft}, a multimodal StarCraft II benchmark supporting both Multi-Agent Reinforcement Learning (MARL) and Vision-Language Model (VLM) paradigms. Unlike SMAC-family environments that rely on abstract state representations and exclude VLMs, AVACraft provides RGB visuals, natural language observations, and structured state information, enabling systematic comparison between training-based and zero-shot methods across 21 scenarios spanning micromanagement, coordination, and strategic planning.
We establish comprehensive baselines: six MARL algorithms (IQL, QMIX, QTRAN, VDN, MAPPO, IPPO) with Swin-Transformer backbones trained for 5M steps, and multiple VLMs including proprietary (GPT-4o) and open-source (Qwen3-VL) models. Results reveal complementary strengths—MARL peaks at 19.3\% win rate after 5M steps, while VLMs achieve 75--90\% zero-shot with human-aligned decisions—exposing trade-offs between training efficiency, performance ceilings, interpretability, and deployment cost. Code: \url{https://github.com/camel-ai/VLM-Play-StarCraft2}.}

\begin{document}
\maketitle

\section{Introduction}

Complex decision-making in dynamic, multi-agent environments represents a fundamental challenge in artificial intelligence, with StarCraft II emerging as a premier testbed due to its real-time nature, partial observability, and requirement for both tactical micromanagement and strategic coordination. Existing StarCraft II benchmarks, including SMAC~\citep{smac} and SMACv2~\citep{smac-v2}, have facilitated significant advances in multi-agent reinforcement learning but suffer from two critical limitations. First, they rely on abstract feature representations that fundamentally diverge from human perception, creating an artificial gap between how AI agents and humans process battlefield information and limiting the ecological validity of learned behaviors. Second, these environments exclusively support traditional reinforcement learning approaches, lacking infrastructure for emerging Vision-Language Models (VLMs) that have demonstrated remarkable zero-shot reasoning capabilities across diverse domains. The rise of foundation models like GPT-4V and Qwen-VL has introduced a new paradigm for AI decision-making that operates without extensive task-specific training, yet their potential in complex, real-time strategic environments remains largely unexplored. While concurrent work such as LLM-PySC2~\citep{llm-pysc2} addresses macro-strategic decision-making and VS-Bench~\citep{vs-bench} evaluates strategic reasoning across multiple games, there remains an urgent need for benchmarks that systematically evaluate and compare both training-based MARL methods and zero-shot VLM approaches specifically for fine-grained tactical micromanagement on equal footing.

We introduce AVACraft, a multimodal StarCraft II benchmark that unifies two interaction paradigms in one framework. It supports RGB visuals, natural language, and structured state inputs, allowing both MARL algorithms and VLM-based agents to be evaluated under the same standardized setting. AVACraft includes 21 scenarios spanning basic control to complex multi-agent coordination, designed to expose the strengths and weaknesses of both approaches. We provide baselines for six MARL methods with Swin-Transformer backbones and multimodal fusion, and several VLMs, including GPT-4o, GPT-4-Turbo, Qwen3-VL-8B, and Qwen3-VL-30B, enabling fair comparison between trained MARL systems and zero-shot VLM agents.

Our experimental evaluation reveals complementary strengths between paradigms: MARL methods achieve peak performance of 19.3\% win rate through extensive training (up to 5M steps) even with state-of-the-art visual backbones, while VLMs demonstrate superior zero-shot capabilities with 75--90\% win rates without any training, producing more interpretable and human-aligned decision processes as validated through expert evaluation involving professional StarCraft II players. The primary contributions of this work include:
\begin{itemize}
    \item We design AVACraft, a multimodal benchmark environment for StarCraft II that supports both MARL and VLM decision-making paradigms through a unified observation space incorporating RGB visual inputs, natural language descriptions, and structured state information. Unlike existing benchmarks that focus on either macro-strategy (LLM-PySC2) or abstract multi-agent settings (VS-Bench), AVACraft targets fine-grained tactical micromanagement with full unit abilities.
    
    \item We establish comprehensive baseline implementations for both paradigms, including six MARL algorithms with Swin-Transformer backbones and multimodal input fusion (trained for 5M steps), and multiple VLMs covering both proprietary and open-source models, along with standardized evaluation protocols, cost analysis, and cross-modal ablation studies.
    
    \item We provide systematic empirical analysis across 21 carefully designed scenarios, revealing fundamental trade-offs between training-based optimization and zero-shot reasoning approaches, supported by expert human evaluation with statistical significance testing demonstrating superior human alignment of VLM-based decisions.
\end{itemize}
Beyond its immediate applications in gaming AI, AVACraft serves as a controlled testbed for studying the intersection of reinforcement learning and foundation models, opening new research directions for developing next-generation AI systems that combine the precision of MARL with the interpretability of VLMs.

\section{AVACraft Benchmark Design}

\begin{figure*}[t]
    \centering
    \includegraphics[width=1\textwidth, trim=0 60 0 60, clip]{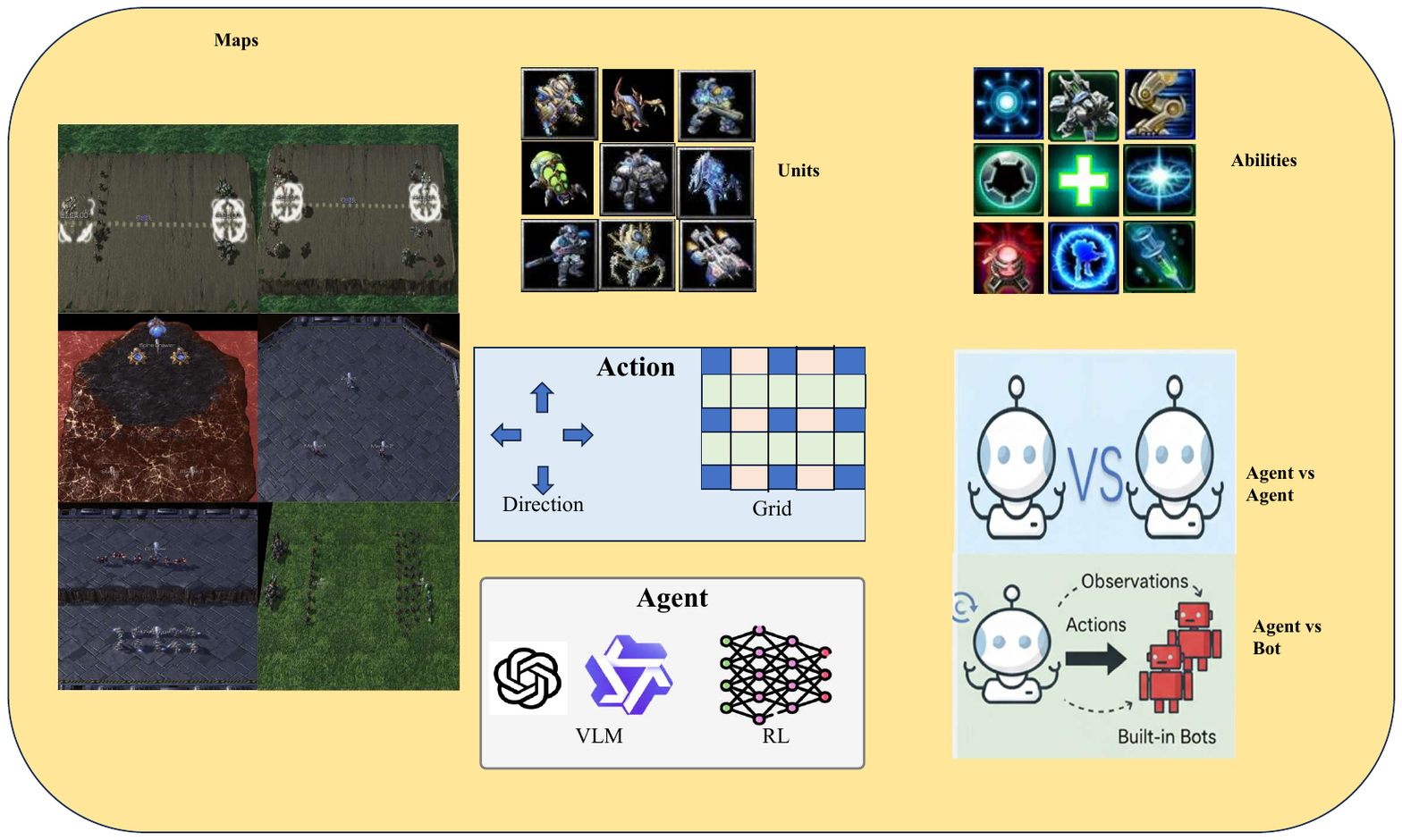}
    \vspace{-4pt}
    \caption{\textbf{AVACraft} environment.}
    \vspace{-10pt}
\end{figure*}

Traditional StarCraft II AI environments like SMAC and SMACv2, while advancing multi-agent reinforcement learning research, suffer from fundamental limitations that hinder the development of comprehensive AI evaluation frameworks. These environments employ abstract feature representations that create substantial perception gaps between AI agents and human players, often modifying unit attributes and employing ``cheat mode'' mechanics that deviate from authentic gameplay. Moreover, they exclusively support reinforcement learning paradigms, lacking the infrastructure necessary for evaluating emerging Vision-Language Models that demonstrate remarkable zero-shot reasoning capabilities. To address these limitations and establish a unified evaluation platform, we introduce AVACraft, a comprehensive multimodal benchmark that supports both traditional MARL approaches and modern VLM-based decision-making within a standardized framework.

\begin{table}[t]
\caption{Comparison of StarCraft II environments. PySC2 provides the foundational API layer.}
\label{tab:env_comparison}
\centering
\small
\setlength{\tabcolsep}{4pt}
\begin{tabular}{@{}lcccc@{}}
\toprule
& \textbf{SMAC} & \textbf{SMACv2} & \textbf{PySC2} & \textbf{AVACraft} \\
\midrule
Visual Input & \xmark & \xmark & Features & \textbf{RGB} \\
Language & \xmark & \xmark & \xmark & \cmark \\
MARL Support & \cmark & \cmark & \xmark & \cmark \\
VLM Support & \xmark & \xmark & \xmark & \cmark \\
Enemy AI & Static & Procedural & Built-in & \textbf{Adaptive} \\
Abilities & Limited & Limited & Full & \textbf{Full} \\
Focus & Algorithms & Generalization & Full game & \textbf{Cross-paradigm} \\
\bottomrule
\end{tabular}
\par
\vspace{1ex}
{\footnotesize \xmark: not supported \quad \cmark: supported \quad \textbf{Bold}: enhanced feature\par}
\vspace{-1em}
\end{table}

AVACraft introduces three key design principles: \textbf{(1) Multi-modal observations} enabling fair comparison between MARL (RGB/scalar) and VLM (RGB+language) approaches; \textbf{(2) Complete unit abilities} preserving StarCraft II's tactical depth unlike SMAC's simplified mechanics; \textbf{(3) Adaptive opponents} preventing strategy exploitation through dynamic policy selection. Table~\ref{tab:env_comparison} summarizes how AVACraft extends beyond existing environments to support cross-paradigm evaluation.

\subsection{Environment Formulation}

We formalize AVACraft as a Partially Observable Markov Decision Process (POMDP) defined by the tuple $\langle \mathcal{S}, \mathcal{A}, \mathcal{O}, P, R, \gamma \rangle$, where $\mathcal{S}$ is the state space, $\mathcal{A}$ is the action space, $\mathcal{O}$ is the observation space, $P: \mathcal{S} \times \mathcal{A} \times \mathcal{S} \rightarrow [0,1]$ is the transition function, $R: \mathcal{S} \times \mathcal{A} \rightarrow \mathbb{R}$ is the reward function, and $\gamma \in [0,1]$ is the discount factor. At each timestep $t$, agents receive partial observations $o_t \in \mathcal{O}$ derived from the true state $s_t \in \mathcal{S}$ and select actions $a_t \in \mathcal{A}$. Note that AVACraft strictly maintains partial observability: the RGB screen and minimap observations natively obey StarCraft II's Fog of War mechanics, so agents only receive visual and textual information within allied units' sight range. The key innovation of AVACraft is providing \textit{flexible observation modes} within $\mathcal{O}$ to support both MARL and VLM paradigms while maintaining a unified evaluation framework.

\subsection{Unified Observation Framework}

The observation space $\mathcal{O}$ provides flexible representations tailored to different AI paradigms while maintaining consistency across evaluation scenarios. We define four observation modes summarized in Table~\ref{tab:obs_modes}.

\begin{table}[h]
\centering
\small
\caption{Observation modes in AVACraft}
\label{tab:obs_modes}
\begin{tabular}{@{}llp{4.5cm}@{}}
\toprule
\textbf{Mode} & \textbf{Notation} & \textbf{Primary Use Case} \\
\midrule
RGB Visual & $o_t^{\text{rgb}}$ & CNN/Transformer-based MARL \\
Scalar Features & $o_t^{\text{feat}}$ & SMAC-compatible MARL \\
Hybrid & $o_t^{\text{hybrid}}$ & Multimodal MARL research \\
VLM-Optimized & $o_t^{\text{vlm}}$ & Vision-Language Models \\
\bottomrule
\end{tabular}
\end{table}

\paragraph{RGB Visual Mode} provides human-like visual observations:
\begin{equation}
    o_t^{\text{rgb}} = (I_t^{\text{scr}}, I_t^{\text{mini}})
\end{equation}
where $I_t^{\text{scr}} \in \mathbb{R}^{H_s \times W_s \times 3}$ captures the main battlefield view (default: $H_s=160, W_s=120$) and $I_t^{\text{mini}} \in \mathbb{R}^{H_m \times W_m \times 3}$ provides tactical overview (default: $H_m=W_m=32$). Resolutions are configurable to balance visual fidelity and computational efficiency.

\paragraph{Scalar Feature Mode} maintains compatibility with existing MARL research through vector representation $o_t^{\text{feat}} \in \mathbb{R}^{d}$ containing unit attributes (health, shields, position, cooldowns) in SMAC-compatible format, enabling direct comparison with prior work.

\paragraph{Hybrid Mode} combines visual and structured information:
\begin{equation}
    o_t^{\text{hybrid}} = (I_t^{\text{scr}}, I_t^{\text{mini}}, o_t^{\text{feat}})
\end{equation}
enabling research into multimodal MARL approaches that leverage both spatial reasoning from images and explicit feature information.

\paragraph{VLM-Optimized Mode} augments visual input with linguistic context:
\begin{equation}
    o_t^{\text{vlm}} = (I_t, T_t, \mathcal{U}_t)
\end{equation}
where:
\begin{itemize}[leftmargin=*,itemsep=1pt]
    \item $I_t$: high-resolution RGB screenshot for visual grounding
    \item $T_t$: natural language description of battlefield state, tactical situation, and mission objectives
    \item $\mathcal{U}_t = \{u_1, \ldots, u_n\}$: structured metadata for each unit where $u_i = (\text{id}_i, \text{type}_i, \text{pos}_i, \text{hp}_i, \text{status}_i)$
\end{itemize}
This multimodal representation lets VLMs use their pre-trained visual reasoning without task-specific training. Natural language descriptions are generated at each timestep to provide precise numerical details, such as HP, shields, and cooldowns, that may be unclear from RGB inputs alone.

\subsection{Action Space Design}
\begin{figure*}[t]
    \centering
    \includegraphics[width=1\textwidth]{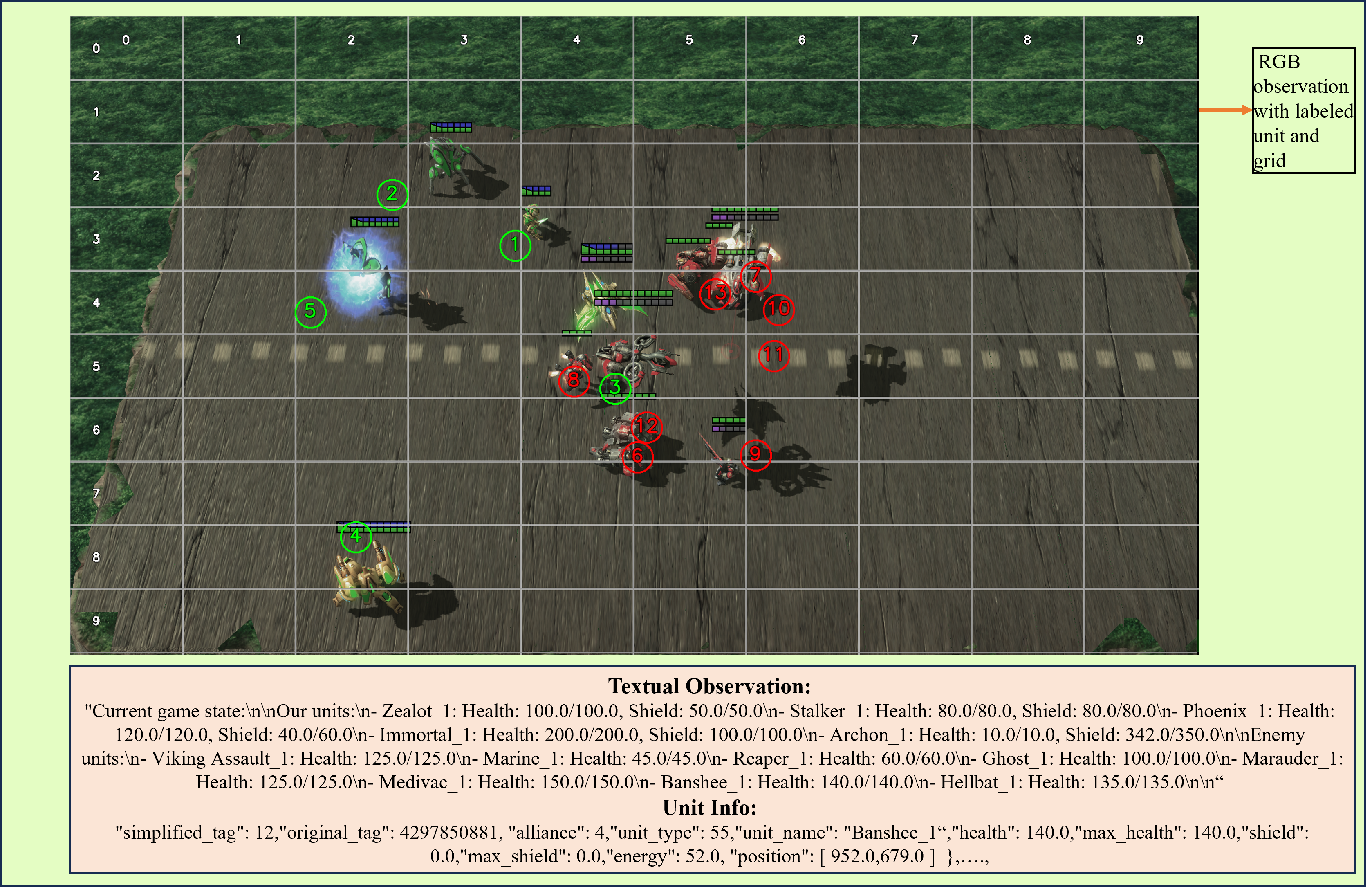}
    \caption{Observation Space of \textbf{AVACraft} environment.}
\end{figure*}
The action space $\mathcal{A}$ supports fine-grained tactical control through three complementary categories:
\begin{equation}
    \mathcal{A} = \mathcal{A}_{\text{atk}} \cup \mathcal{A}_{\text{mov}} \cup \mathcal{A}_{\text{abl}}
\end{equation}

\textbf{Attack Actions} ($\mathcal{A}_{\text{atk}}$): Ordered pairs $(i,j)$ specifying unit $i$ targeting enemy unit $j$, enabling focus-fire and target prioritization strategies critical for tactical micromanagement.

\textbf{Movement Actions} ($\mathcal{A}_{\text{mov}}$): Support both precise and directional positioning:
\begin{itemize}[leftmargin=*,itemsep=2pt]
    \item \textit{Grid positioning}: $(i, x, y)$ where $(x,y) \in [1,10]^2$ provides discrete spatial coordinates for battlefield coordination
    \item \textit{Directional movement}: $(i, d)$ where $d \in \{\text{UP, DOWN, LEFT, RIGHT}\}$ enables rapid repositioning like SMAC
\end{itemize}

\textbf{Ability Actions} ($\mathcal{A}_{\text{abl}}$): Triples $(i, \text{ability}, \text{target})$ covering key StarCraft II tactics, including defensive, offensive, and mobility abilities. Unlike SMAC, AVACraft retains unit abilities, preserving the tactical richness of high-level gameplay. Targets may be positions, unit IDs, or null, depending on the ability.

\subsection{Adaptive Enemy Policies and Standardized Evaluation}

To ensure robust evaluation across different challenge levels, AVACraft implements an adaptive enemy system extending beyond traditional static AI opponents. Drawing inspiration from SMAC-Hard~\citep{smac-hard}, we develop a multi-tier enemy policy framework:

\textbf{Built-in AI}: StarCraft II difficulty level 7 (VeryHard) provides a consistent and competent baseline opponent.

\textbf{Script-based Policies}: Three specialized behavior policies per scenario generated through LLM-assisted behavior tree synthesis, each emphasizing different tactical approaches (aggressive rushing, defensive positioning, economic optimization). This diversity prevents agents from overfitting to single strategies.

\textbf{Randomized Selection}: Dynamic policy selection during evaluation ensures generalization assessment by preventing exploitation of predictable opponent patterns.

Our benchmark encompasses 21 carefully designed scenarios spanning multiple complexity dimensions: 12 core micromanagement scenarios testing fundamental tactical skills (unit control, target prioritization, ability timing), 5 coordination scenarios requiring multi-unit synchronization and formation control, and 4 strategic scenarios incorporating terrain utilization and resource management. Each scenario supports both PvE and PvP modes, with PvE scenarios featuring the adaptive enemy system and PvP scenarios enabling direct agent-versus-agent competition for comparative evaluation between paradigms. Detailed scenario specifications are provided in Appendix~\ref{map_details}.

AVACraft employs a sparse reward structure $R(s_t) \in \{-1, 0, 1\}$ corresponding to defeat, ongoing/draw, and victory states respectively. Episodes terminate under three conditions: complete enemy elimination (victory), complete allied elimination (defeat), or 300-second time limit (draw). This design provides clear performance signals while maintaining tactical flexibility and avoiding reward shaping that might bias particular approaches.

\section{Baseline Implementations}

To establish comprehensive baselines for both paradigms supported by AVACraft, we implement representative approaches for Vision-Language Model agents and Multi-Agent Reinforcement Learning algorithms. These baselines serve as reference implementations for the research community and demonstrate the benchmark's capability to fairly evaluate fundamentally different decision-making approaches.

\subsection{VLM-based Decision Making: Attentive VLM Agent (AVA)}

We develop AVA as our primary VLM baseline, designed to leverage the multimodal reasoning capabilities of foundation models for strategic decision-making. The AVA architecture integrates three key components: a Multimodal Priority Inference mechanism for strategic unit targeting, a knowledge-enhanced decision system through retrieval-augmented generation, and a dynamic role assignment framework for coordinated multi-agent behavior. We emphasize that AVA is a proof-of-concept baseline that validates the environment's capabilities for evaluating VLM-based agents, rather than a novel architectural contribution.

\subsubsection{Multimodal Priority Inference Mechanism}
Our priority inference system processes battlefield information through structured skill planning and tactical decision-making. The mechanism operates in two key stages to identify and prioritize strategic elements. First, we implement a VLM Planner that evaluates the battlefield situation and generates specific micro-management skill plans:
\begin{equation}
    S = \text{VLM}_{\text{plan}}(I, T, H) = \{s_{\text{primary}}, s_{\text{secondary}}\},
\end{equation}
where the planner outputs structured skill plans with primary and secondary tactical objectives. Based on the planner's output, the system performs precise unit identification and classification:
\begin{equation}
    A = \text{VLM}_{\text{detect}}(I) = \{a_1, ..., a_n\},
\end{equation}
where each annotation $a_i = (p_i, c_i, b_i)$ consists of unit position $p_i$, unit class $c_i$, and bounding box $b_i$ for accurate spatial localization.

The critical Multimodal Priority Inference process then integrates visual features with tactical objectives through skill-aware natural language prompting:
\begin{equation}
    U_{\text{priority}} = \text{VLM}_{\text{analyze}}(I, T, H, A, Q, S),
\end{equation}
where $I$ is the current game screenshot, $T$ is the text state description, $H$ represents action history for temporal reasoning, $A$ is the set of unit annotations, $Q$ is the tactical analysis prompt generated based on the skill plan, and $S$ is the current skill plan from the VLM Planner. The VLM outputs its analysis in structured natural language, integrating battlefield assessment with tactical prioritization.

\subsubsection{Knowledge Integration through RAG}
To enhance tactical decision-making with domain expertise, we implement a Retrieval-Augmented Generation system that operates on priority units identified through Multimodal Priority Inference. Given the priority unit set $U_{\text{priority}} \subseteq A$, we formulate the knowledge retrieval and integration process as:
\begin{equation}
    K(u) = \text{Retrieve}(c_u) = \{s_u, m_u, t_u\} \quad \forall u \in U_{\text{priority}},
\end{equation}
where for each unit $u$ with class $c_u$, we retrieve a knowledge tuple $K(u)$ consisting of unit specifications $s_u$, matchup data $m_u$, and tactical insights $t_u$. The retrieved knowledge is then integrated with the current game state through a context-aware generation process:
\begin{equation}
    D = \text{VLM}_{\text{synthesize}}(I, T, H, U_{\text{priority}}, \{K(u)\}),
\end{equation}
where $D$ represents the tactical decision guidance generated by combining retrieved knowledge with current game state representation.

\subsubsection{Dynamic Role Assignment and Decision Pipeline}
For multi-agent coordination, we implement a dynamic role assignment framework that adapts to evolving battlefield conditions. Let $\mathcal{N} = \{1, ..., N\}$ denote the set of agents and $\mathcal{Z} = \{z_1, ..., z_m\}$ represent possible roles. The role assignment function $\phi: \mathcal{N} \rightarrow \mathcal{Z}$ maps each agent to a specific role, optimized through a utility function $U(\phi, s)$ that evaluates role effectiveness given the current state $s \in \mathcal{S}$. Our framework leverages VLMs through a multimodal fusion function $z_i=\text{VLM}_\text{role}(I,T,C)$, where the model processes visual inputs $I$, textual prompts $T$, and contextual information $C$ to generate role assignments.

The complete decision-making pipeline maps POMDP observations to actions through VLM-based transformations. At each timestep $t$, given observation $o_t = (I_t, T_t, U_t)$ from AVACraft environment state $s_t \in \mathcal{S}$, our system generates actions by maintaining a history buffer $H_t$ and processing each step to maximize the trade-off between strategic depth and real-time responsiveness.

\subsection{MARL-based Decision Making}

For the MARL paradigm, we implement six representative algorithms spanning both value-decomposition and policy-gradient methods: Independent Q-Learning (IQL), QMIX, QTRAN, Value Decomposition Networks (VDN), Multi-Agent PPO (MAPPO)~\citep{mappo}, and Independent PPO (IPPO). These algorithms are adapted to work with AVACraft's visual observations through state-of-the-art visual processing architectures.

Our upgraded MARL implementation employs a \textbf{Swin-Transformer} (Swin-Tiny, 27.5M parameters) as the visual backbone, replacing the simple CNN used in preliminary experiments to ensure that RL agents have access to competitive visual feature extraction. The visual encoder processes both screen ($160 \times 120$) and minimap ($32 \times 32$) observations through separate Swin-Tiny streams with adaptive pooling to produce fixed-size representations.

We also introduce a \textbf{multimodal fusion} mode for MARL, combining visual features with pre-computed text embeddings. Natural language observations are encoded by GTE-Base and fused with visual features through a learned projection layer before entering the mixing networks, enabling direct comparison with VLMs using the same textual information.

The algorithms differ in their coordination mechanisms: IQL treats agents independently, QMIX uses a monotonic mixing network, QTRAN relaxes the monotonicity constraint, VDN sums individual Q-values, while MAPPO and IPPO use centralized and independent critic architectures respectively with PPO policy updates. All implementations support the observation modes provided by AVACraft (RGB visual, SMAC-compatible scalar, hybrid, and vision+text), enabling comparative analysis of how different input modalities affect learning efficiency and final performance.

\section{Experimental Evaluation}

We conduct comprehensive experiments to evaluate both VLM and MARL paradigms within the AVACraft benchmark, focusing on four primary objectives: (1) establishing performance baselines for both paradigms across diverse scenarios, (2) conducting systematic cross-paradigm comparison with cross-modal ablations, (3) validating human alignment through expert evaluation with statistical testing, and (4) analyzing computational costs and scalability. Our experimental setup leverages dual A100 40GB GPUs for MARL training and the Camel framework for VLM agent coordination, with all experiments conducted at 2Hz frequency to balance strategic decision depth and computational efficiency. Detailed hyperparameters are provided in Appendix~\ref{app:hyperparameters}.

\subsection{Cross-Paradigm Performance Analysis}

We evaluate both paradigms across a representative subset of AVACraft scenarios. For MARL evaluation, we implement six algorithms using RGB visual observations processed through Swin-Tiny backbones and train for \textbf{5 million steps} on the foundational 3m scenario, following standard SMAC practice. We evaluate MARL agents under both Vision-Only and Vision+Text input modes. For VLM evaluation, we assess multiple models including proprietary (GPT-4-Turbo, GPT-4o, GPT-4o-mini) and open-source (Qwen-VL-Plus, Qwen3-VL-8B, Qwen3-VL-30B) models across 12 micromanagement scenarios.

\begin{table}[t]
\caption{Cross-paradigm performance comparison on the 3m scenario. MARL results after 5M training steps with Swin-Tiny backbone. Win rates reported as mean $\pm$ std over 5 seeds. \textsuperscript{$\dagger$}Vision+Text mode uses GTE-Base text embeddings fused with visual features.}
\label{tab:cross_paradigm_3m}
\centering
\small
\begin{tabular}{@{}llcc@{}}
\toprule
\textbf{Method} & \textbf{Input Mode} & \textbf{Steps} & \textbf{Win Rate (\%)} \\
\midrule
\multicolumn{4}{l}{\textit{MARL Methods (Swin-Tiny backbone)}} \\
MAPPO & Vision+Text\textsuperscript{$\dagger$} & 5M & 19.3 $\pm$ 3.2 \\
IPPO & Vision Only & 5M & 18.2 $\pm$ 2.8 \\
IPPO & Vision+Text\textsuperscript{$\dagger$} & 5M & 16.6 $\pm$ 3.5 \\
QMIX & Vision Only & 5M & 27.1 $\pm$ 4.1 \\
QTRAN & Vision Only & 5M & 2.0 $\pm$ 1.4 \\
IQL & Vision Only & 5M & 0.0 $\pm$ 0.0 \\
VDN & Vision Only & 5M & 0.0 $\pm$ 0.0 \\
\midrule
\multicolumn{4}{l}{\textit{VLM Methods (Zero-shot, proprietary)}} \\
GPT-4o & VLM-Optimized & 0 & \textbf{81} $\pm$ 3.9 \\
GPT-4-Turbo & VLM-Optimized & 0 & 79 $\pm$ 4.1 \\
GPT-4o-mini & VLM-Optimized & 0 & 76 $\pm$ 4.3 \\
Qwen-VL-Plus & VLM-Optimized & 0 & 75 $\pm$ 4.3 \\
\midrule
\multicolumn{4}{l}{\textit{VLM Methods (Zero-shot, open-source)}} \\
Qwen3-VL-30B & VLM-Optimized & 0 & 50 $\pm$ 5.0 \\
Qwen3-VL-8B & VLM-Optimized & 0 & 40 $\pm$ 4.9 \\
\bottomrule
\vspace{-1.5em}
\end{tabular}
\end{table}

Table~\ref{tab:cross_paradigm_3m} reveals striking differences between paradigms on the foundational 3m scenario. Even with upgraded Swin-Transformer backbones and 5M training steps, MARL methods achieve at most 27.1\% win rate (QMIX), with policy-gradient methods (MAPPO, IPPO) plateauing around 16--19\%. In contrast, VLM approaches demonstrate superior zero-shot capabilities, with proprietary models achieving 75--81\% win rates and open-source Qwen3-VL models achieving 40--50\% win rates without any training, highlighting the power of pre-trained foundation models for strategic reasoning.

\paragraph{Cross-Modal Ablation.} An important finding emerges from the MARL cross-modal ablation: IPPO with Vision+Text input (16.6\%) slightly underperforms IPPO with Vision-Only input (18.2\%), suggesting that from-scratch MARL agents struggle to effectively fuse pre-computed text embeddings with visual features during training. In contrast, VLMs inherently align text and image modalities through pre-training, naturally leveraging the natural language channel that provides exact numerical values (HP, cooldowns, unit IDs) critical for precise tactical decisions. This ablation demonstrates AVACraft's utility in exposing the unique cross-modal alignment capabilities of foundation models versus traditional RL architectures.

\begin{table}[t]
\caption{VLM performance across AVACraft scenarios. Win rates (\%) for zero-shot evaluation (mean $\pm$ std over 20 episodes).}
\label{tab:vlm_comprehensive}
\centering
\small
\begin{tabular}{@{}lrrrrl@{}}
\toprule
\textbf{Scenario} & \textbf{GPT-4o} & \textbf{Qwen-VL} & \textbf{Q3-30B} & \textbf{Q3-8B} & \textbf{Challenge} \\
\midrule
\multicolumn{6}{l}{\textit{Low Complexity}} \\
3m & \textbf{81} & 75 & 50 & 40 & Coordination \\
2m\_vs\_1z & \textbf{23} & 10 & 15 & 5 & Micro control \\
\midrule
\multicolumn{6}{l}{\textit{Medium Complexity}} \\
mixed\_units & \textbf{79} & 75 & 60 & 35 & Targeting \\
2s3z & \textbf{41} & 25 & 20 & 10 & Unit synergy \\
3s\_vs\_3z & \textbf{32} & 10 & 15 & 5 & Positioning \\
2s\_vs\_1sc & \textbf{5} & 0 & 0 & 0 & Range mgmt. \\
\midrule
\multicolumn{6}{l}{\textit{High Complexity}} \\
pvz\_ht & \textbf{34} & 25 & 20 & 10 & Ability timing \\
8m2st\_vs\_35zg4b & \textbf{53} & 25 & 30 & 15 & Formation \\
8m1mv\_vs\_2st & \textbf{12} & 0 & 5 & 0 & Support coord. \\
8m\_vs\_2pc1wp & \textbf{11} & 0 & 5 & 0 & Terrain \\
6r\_vs\_8z & 0 & 0 & 0 & 0 & Hit-and-run \\
\midrule
\multicolumn{6}{l}{\textit{Very High Complexity}} \\
2c\_vs\_64zg & 0 & 0 & 0 & 0 & AOE optim. \\
\midrule
\textbf{Average} & \textbf{30.9} & 20.4 & 18.3 & 10.0 & \\
\bottomrule
\end{tabular}
\vspace{-1.5em}
\end{table}

Table~\ref{tab:vlm_comprehensive} shows that both proprietary and open-source VLMs perform well on low- to medium-complexity scenarios, but struggle on harder ones. GPT-4o and Qwen-VL achieve strong results on tasks like mixed\_units and 3m, while Qwen3-VL-30B is competitive on simpler settings, supporting reproducible evaluation without proprietary APIs. However, all models fail on the most difficult scenarios, with 0\% win rates on 2c\_vs\_64zg and 6r\_vs\_8z.

\paragraph{Error Analysis of 0\% Win Rate Scenarios.} The universal 0\% win rate on 2c\_vs\_64zg and 6r\_vs\_8z across all VLMs—including Qwen3-VL-30B which achieves 90\% on simpler maps—confirms that these failures reflect the true capability ceiling of current VLMs for dense spatial reasoning and high-frequency micro-control, rather than environment bugs or prompt engineering issues. Specifically, 2c\_vs\_64zg requires precise AOE line-damage optimization against 64 units with continuous kiting, while 6r\_vs\_8z demands sustained hit-and-run micro over many timesteps. Both require spatial precision and temporal consistency that exceed current VLM capabilities.

\subsection{Architectural Component Analysis}

To understand the contribution of different components in our AVA baseline, we conduct a comprehensive ablation study using GPT-4-Turbo on the mixed\_units scenario. The three key components are: (1) Dynamic Role Assignment for coordination, (2) Multimodal Priority Inference (MPI) for target selection, and (3) Retrieval-Augmented Generation (RAG) for domain knowledge integration.

\begin{table}[H]
\caption{Component ablation study showing individual and combined contributions of AVA architecture components.}
\label{tab:ablation_comprehensive}
\centering
\small
\begin{tabular}{@{}cccc@{}}
\toprule
\textbf{Role} & \textbf{MPI} & \textbf{RAG} & \textbf{Win Rate (\%)} \\
\midrule
\checkmark & \checkmark & \checkmark & \textbf{87} $\pm$ 3.4 \\
\checkmark & \checkmark & - & 71 $\pm$ 4.5 \\
\checkmark & - & \checkmark & 65 $\pm$ 4.8 \\
- & \checkmark & \checkmark & 70 $\pm$ 4.6 \\
\checkmark & -  & - & 24 $\pm$ 4.3 \\
- & \checkmark & - & 50 $\pm$ 5.0 \\
- & - & \checkmark & 26 $\pm$ 4.4 \\
- & - & - & 20 $\pm$ 4.0 \\
\bottomrule
\end{tabular}
\end{table}

\begin{table}[H]
\caption{Head-to-head VLM comparison on mixed\_units scenario (20 matches per pairing).}
\label{tab:model_vs_model_compact}
\centering
\small
\begin{tabular}{@{}lccccc@{}}
\toprule
\textbf{Model} & \textbf{GPT-4o} & \textbf{GPT-4T} & \textbf{Qwen} & \textbf{Gemini-F} & \textbf{Win Rate} \\
\midrule
GPT-4o & -- & 9:11 & 13:7 & 9:11 & 55\% \\
GPT-4-Turbo & 11:9 & -- & 14:6 & 8:12 & 58\% \\
Qwen-VL & 7:13 & 6:14 & -- & 8:12 & 35\% \\
Gemini-Flash & 11:9 & 12:8 & 12:8 & -- & 62\% \\
\bottomrule
\end{tabular}
\end{table}

The ablation results (Table~\ref{tab:ablation_comprehensive}) demonstrate the complementary nature of AVA's components. The complete system achieves 87\% win rate, with MPI providing the most substantial individual contribution (50\% vs 20\% baseline), RAG contributing 20--25\% improvement through domain knowledge, and Role Assignment adding 15--20\% through enhanced coordination. Notably, all components show positive interactions, with the combined system significantly outperforming any individual component.

\subsection{Human Alignment Evaluation}

To assess the human-like qualities of decision-making across paradigms, we conduct a structured evaluation with seven participants representing diverse StarCraft II expertise levels: one professional player, two Master-level players, one Diamond-level player, one Platinum-level player, one Gold-level player, one novice, and one spectator. Participants evaluate both VLM (AVA) and MARL agents across three metrics on a 1--5 scale: Game Bug Exploitation (higher indicating less exploitation), Reasoning Coherence, and Human Similarity. Evaluations were conducted in a blinded setting where participants were not informed which agent type they were evaluating. See Appendix~\ref{app:evaluation_metrics} for detailed metric definitions.

\begin{table}[t]
\caption{Human evaluation comparing VLM and MARL approaches. Scores: 1--5 scale (higher is better). Statistical significance tested via Mann-Whitney U test.}
\label{tab:human_evaluation_comprehensive}
\centering
\small
\begin{tabular}{@{}lccc@{}}
\toprule
\textbf{Evaluator Group} & \textbf{Metric} & \textbf{MARL} & \textbf{VLM} \\
\midrule
\multirow{3}{*}{\textbf{Expert} (n=3)} 
& Bug Exploit.$^*$ & 1.3 & \textbf{5.0} \\
& Reasoning & 2.0 & \textbf{4.3} \\
& Human Sim. & 1.0 & \textbf{4.7} \\
\midrule
\multirow{3}{*}{\textbf{Mid-tier} (n=3)}
& Bug Exploit.$^*$ & 2.0 & \textbf{3.7} \\
& Reasoning & 2.0 & \textbf{4.7} \\
& Human Sim. & 1.3 & \textbf{4.7} \\
\midrule
\multirow{3}{*}{\textbf{Novice} (n=2)}
& Bug Exploit.$^*$ & 2.5 & \textbf{4.0} \\
& Reasoning & 2.5 & \textbf{3.5} \\
& Human Sim. & 3.0 & \textbf{4.0} \\
\midrule
\textbf{Overall Average} & & 1.9 & \textbf{4.4}$^{\dagger\dagger}$ \\
\bottomrule
\end{tabular}
\vspace{0.5ex}

\footnotesize{$^*$Higher = less bug exploitation. $^{\dagger\dagger}$Mann-Whitney $U$ test: $p < 0.01$ for all three metrics (VLM vs MARL overall).}
\end{table}

Table~\ref{tab:human_evaluation_comprehensive} shows VLM agents significantly outperforming MARL approaches across all metrics and expertise levels ($p < 0.01$, Mann-Whitney $U$ test). Expert evaluators were particularly critical of MARL agents, noting frequent exploitation of environment mechanics and poor strategic coherence (average scores of 1.3--2.0). In contrast, VLM agents received consistently high ratings (4.3--4.5 average) for producing interpretable, human-like decision processes. Expert evaluators specifically highlighted VLM agents' implementation of advanced tactical principles including armor-type targeting, focus-fire coordination, and formation control that closely resemble professional gameplay strategies.

\subsection{Computational Cost and Scalability Analysis}

We analyze the computational requirements and deployment costs of both paradigms to provide practical guidance for researchers and practitioners.

\begin{table}[t]
\caption{Computational cost comparison between paradigms. MARL costs reflect 5M-step training on dual A100 40GB GPUs. VLM costs reflect per-episode inference on the 3m scenario (avg.\ 150 steps/episode).}
\label{tab:cost_analysis}
\centering
\small
\begin{tabular}{@{}lcccc@{}}
\toprule
\textbf{Method} & \textbf{Training} & \textbf{Latency} & \textbf{Tokens/} & \textbf{Cost/} \\
& \textbf{Time} & \textbf{(s/step)} & \textbf{Decision} & \textbf{Episode} \\
\midrule
\multicolumn{5}{l}{\textit{MARL Methods}} \\
MAPPO (Swin-T) & $\sim$72h & 0.008 & -- & GPU only \\
IPPO (Swin-T) & $\sim$65h & 0.008 & -- & GPU only \\
QMIX (Swin-T) & $\sim$80h & 0.010 & -- & GPU only \\
\midrule
\multicolumn{5}{l}{\textit{VLM Methods (Proprietary)}} \\
GPT-4o & 0 & 2.3 & $\sim$1,800 & $\sim$\$4.05 \\
GPT-4-Turbo & 0 & 3.1 & $\sim$2,100 & $\sim$\$6.30 \\
GPT-4o-mini & 0 & 1.2 & $\sim$1,500 & $\sim$\$0.45 \\
\midrule
\multicolumn{5}{l}{\textit{VLM Methods (Open-source, single A100)}} \\
Qwen3-VL-30B & 0 & 4.5 & $\sim$1,600 & $\sim$\$0 \\
Qwen3-VL-8B & 0 & 1.8 & $\sim$1,400 & $\sim$\$0 \\
\bottomrule
\end{tabular}
\end{table}

Table~\ref{tab:cost_analysis} reveals fundamental cost trade-offs between paradigms. MARL training requires approximately 65--80 hours on dual A100 GPUs for 5M steps on the 3m scenario, with significant memory requirements for experience replay buffers. Once trained, MARL agents achieve fast inference ($\sim$8--10ms per step). VLM agents operate with zero training overhead but incur per-decision costs: GPT-4o averages 2.3 seconds and $\sim$\$0.027 per decision, resulting in $\sim$\$4.05 per episode. Open-source Qwen3-VL models eliminate API costs entirely and can be deployed on a single A100 GPU, making fully reproducible evaluation accessible without proprietary dependencies. For scenarios requiring rapid deployment or frequent scenario changes, VLM approaches offer significant advantages, while MARL methods may be preferred for high-frequency, cost-sensitive applications once training is completed.

\subsection{Key Findings and Implications}

Our comprehensive evaluation reveals several critical insights about AI decision-making paradigms in complex strategy environments:

\textbf{Zero-shot vs Training Trade-off}: VLM agents demonstrate remarkable zero-shot capabilities, achieving 75--81\% win rates on fundamental scenarios without any training, while MARL methods struggle to achieve comparable performance even after 5M training steps with state-of-the-art Swin-Transformer backbones, confirming the extreme sample inefficiency of training spatial/tactical policies from scratch.

\textbf{Cross-Modal Alignment Gap}: MARL agents fail to leverage textual observations effectively (IPPO Vision+Text underperforms Vision-Only), while VLMs naturally exploit cross-modal information through pre-trained alignment. This gap highlights the unique advantage of foundation model pre-training for multimodal tactical reasoning.

\textbf{Complexity Limitations}: Both paradigms show performance degradation on high-complexity scenarios, but through different failure modes. MARL agents fail to learn effective coordination strategies from visual inputs, while VLM agents struggle with precise timing and micro-management despite strong strategic understanding. The 0\% win rates on the most challenging scenarios across all VLMs represent genuine capability ceilings rather than engineering failures.

\textbf{Human Alignment}: VLM agents produce significantly more interpretable and human-like decision processes ($p < 0.01$), making them valuable for applications requiring explainable AI or human-AI collaboration.

\textbf{Cost-Performance Landscape}: Open-source VLMs (Qwen3-VL-30B) achieve competitive performance on simpler scenarios at zero marginal cost, while proprietary models maintain advantages on complex tasks. MARL offers fast inference after expensive training, creating complementary deployment profiles.

\section{Conclusion}
AVACraft establishes a standardized multimodal benchmark for cross-paradigm evaluation in StarCraft II, enabling direct comparison between training-based MARL and zero-shot VLM approaches via unified RGB, language, and structured observations. Comprehensive evaluation across 21 scenarios with six MARL algorithms and multiple VLMs (proprietary and open-source) reveals fundamental trade-offs in sample efficiency, cross-modal reasoning, interpretability, and deployment cost—MARL peaks at 19.3\% after 5M steps, while VLMs achieve 75--90\% zero-shot with human-aligned decisions. Beyond StarCraft II, AVACraft offers a framework for studying human-aligned AI, opening future directions including hybrid RL-VLM systems, enhanced spatial reasoning for dense formations, and scaling to full-game scenarios.

\section*{Limitations}
AVACraft focuses on tactical micromanagement rather than full-game long-horizon tasks with resource and tech-tree management, and VLMs achieve 0\% win rates on the most challenging maps requiring precise spatial reasoning and sustained micro-control. MARL agents may benefit from longer training budgets or curriculum strategies beyond our 5M-step evaluation. The human study, while statistically significant ($p < 0.01$), involves only seven participants. Current metrics center on win rate; finer-grained measures (APM, micro-action accuracy, token efficiency) and hybrid RL-VLM frameworks are promising directions we leave to future work.

\newpage

\bibliographystyle{acl_natbib}
\bibliography{references}

\appendix
\clearpage

\section{Impact Statement}
This work advances the field of multimodal AI decision-making through the lens of real-time strategy games. While our primary contribution is methodological, we acknowledge several potential societal implications. The development of more human-aligned AI agents could enhance human-AI collaboration and improve AI system interpretability. However, advances in strategic decision-making capabilities also warrant careful consideration regarding dual-use applications. We believe our focus on human-centric design and transparent decision processes helps promote responsible AI development. Our framework primarily serves as a research tool for studying AI capabilities in controlled game environments, with minimal risk of direct negative societal impact.

\section{Related Work}

\paragraph{Foundation Models for Multimodal Understanding}
Recent advances in Large Language Models such as GPT-4~\citep{gpt4}, Claude~\citep{claude}, and Llama~\citep{llama} have demonstrated remarkable reasoning capabilities. Building upon these foundations, Vision-Language Models~\citep{clip,llava} integrate visual encoders with language models, enabling simultaneous understanding of visual and textual information. Models including GPT-4V~\citep{gpt4v}, Gemini~\citep{gemini}, and Qwen-VL~\citep{qwenvl} have shown strong performance across diverse multimodal tasks, with applications spanning robotic control~\citep{rt2}, web navigation~\citep{webvoyager}, and interactive environments~\citep{cradle}.

\paragraph{Vision-Language Models for Game Environments}
Game environments have emerged as important testbeds for evaluating VLM decision-making capabilities. CRADLE~\citep{cradle} introduced the General Computer Control framework, demonstrating that VLMs can interact with complex AAA games like Red Dead Redemption 2 using only screenshots and keyboard-mouse actions. Minecraft has become a particularly popular platform for VLM agent research. The STEVE series~\citep{steve} combines vision models with LLMs for embodied agents capable of open-world exploration. GROOT~\citep{groot} learns instruction following by watching gameplay videos without manual annotations. JARVIS-VLA~\citep{jarvis} employs vision-language post-training for end-to-end action prediction. MCU Benchmark~\citep{mcu} provides a systematic evaluation framework with 3,452 atomic tasks spanning diverse skills including manipulation, navigation, and combat. Cross-game benchmarks have also been developed: BALROG~\citep{balrog} aggregates six RL environments including BabyAI, Crafter, and NetHack to evaluate long-horizon decision-making capabilities across different game genres. VideoGameBench~\citep{videogamebench} includes 23 classic games requiring VLMs to complete entire games using only raw visual inputs, providing insights into VLM capabilities across varied gameplay mechanics. VS-Bench~\citep{vs-bench} evaluates VLMs for strategic reasoning and decision-making in multi-agent environments across multiple games, focusing on high-level strategic capabilities. These works have demonstrated VLMs' potential for understanding game environments and generating appropriate actions based on visual observations.

\paragraph{StarCraft II as an AI Benchmark}
StarCraft II has served as a premier benchmark for artificial intelligence research, particularly for multi-agent systems requiring real-time coordination under partial observability. AlphaStar~\citep{alphastarnature} achieved superhuman performance through a combination of imitation learning from human replays and multi-agent reinforcement learning, demonstrating that deep RL could master the game's full complexity. This work inspired numerous architectural improvements including distributed training frameworks~\citep{starcraft2unplugged}, hierarchical decision-making~\citep{distar}, and macro-action abstractions~\citep{Tstarbot-x,ROA-Star}. For standardized multi-agent evaluation, the StarCraft Multi-Agent Challenge (SMAC)~\citep{smac} provided a widely-adopted framework focusing on cooperative micromanagement scenarios with decentralized execution. SMAC has facilitated significant advances in value decomposition methods, communication protocols, and credit assignment mechanisms. SMACv2~\citep{smac-v2} extended this foundation by introducing procedurally generated scenarios requiring adaptive closed-loop policies rather than exploiting fixed opponent behaviors. SMAC-Hard~\citep{smac-hard} further increased tactical complexity through scenarios demanding precise ability usage and unit coordination. These benchmarks have collectively advanced multi-agent reinforcement learning research through standardized evaluation protocols and diverse tactical challenges.

\paragraph{Language Models for StarCraft II Decision-Making}
Recent works have begun exploring the integration of language models with StarCraft II environments. LLM Play SC2~\citep{llm-play-sc2} pioneered the application of LLMs to macro-strategic decision-making in full matches, developing the TextStarCraft II text-based environment that enables LLMs to make high-level decisions regarding resource management, unit production, and technology progression. LLM-PySC2~\citep{llm-pysc2} provides comprehensive access to the complete PySC2 action space along with multimodal observation interfaces including visual inputs, minimap information, and structured game state. The framework includes built-in game knowledge and example demonstrations to facilitate LLM understanding of game mechanics. LLM-SMAC~\citep{llm-smac} demonstrates the potential of code generation paradigms for tactical decision-making by leveraging LLMs to generate decision tree code for SMAC scenarios, enabling interpretable policy representation. Additional works~\citep{Advancing_DRL,HIFAS} have explored learning from language-based strategy descriptions and hierarchical planning. These approaches have shown that language models can understand StarCraft II's strategic and tactical concepts through textual descriptions and code generation.

While existing work has advanced both VLM-based game AI and StarCraft II research independently, current benchmarks lack unified evaluation frameworks that support both traditional MARL and modern VLM approaches for fine-grained tactical micromanagement. SMAC-family benchmarks employ abstract state representations incompatible with VLM perception, while VLM game research has primarily focused on macro-level strategies (LLM-PySC2) or cross-game evaluation (VS-Bench) rather than precise multi-unit coordination. AVACraft addresses this gap by providing multimodal observations—RGB visuals, natural language descriptions, and structured state information—within a standardized evaluation framework, enabling systematic comparison between training-based and zero-shot decision-making paradigms in tactical control scenarios.

\section{Limitations of Previous StarCraft II Environments}
\label{app:smac_limitations}

While SMAC and SMACv2 have advanced multi-agent reinforcement learning research, they have fundamental limitations for developing AI systems that can truly master StarCraft II's complex decision-making challenges:

\paragraph{Simplified Unit Abilities and Interactions} 
SMAC significantly simplifies unit abilities, removing critical micro-management elements that define StarCraft II gameplay. For example, Marines and Marauders lack Stimpack abilities, Stalkers cannot Blink, and only Medivacs retain their Heal ability. This oversimplification eliminates the rich tactical depth of StarCraft II, where ability timing and targeting often determine battle outcomes. In competitive play, a Marine without Stimpack is essentially a different unit, and skilled micro-management of these abilities is central to high-level play.

\paragraph{Limited Unit Diversity and Compositions}
Both SMAC and SMACv2 feature extremely limited unit diversity, with most scenarios containing only 2-3 unit types. This fails to capture StarCraft II's emphasis on complementary unit compositions and counter strategies. For instance, the classic ``Marine-Marauder-Medivac'' composition requires specific control patterns that balance front-line positioning, focus fire, and healing priorities—tactical considerations absent in simplified environments.

\paragraph{Overly Simple Enemy AI} 
The enemy AI in SMAC and SMACv2 follows a basic ``attack spawn point'' strategy without any tactical depth. It neither repositions units strategically nor prioritizes targets intelligently, creating unrealistic combat scenarios. This simplistic behavior fails to challenge agents to develop the sophisticated positioning and targeting skills needed in actual StarCraft II gameplay, resulting in strategies that don't transfer to real matches.

\paragraph{Abstract State Representations}
SMAC and SMACv2 represent the game state as abstract vectors containing unit attributes, positions, and health values, completely divorced from the visual and spatial reasoning humans use when playing. This misalignment between AI and human perception fundamentally limits the ecological validity of behaviors learned in these environments.

\paragraph{Questionable Randomization in SMACv2}
While SMACv2 introduces procedural generation and randomization of unit types and positions, these changes don't necessarily reflect meaningful tactical variations in StarCraft II. Random army compositions often create unrealistic scenarios that wouldn't occur in competitive play, where army composition follows strategic principles and tech progression. This randomization tests an agent's ability to handle arbitrary unit combinations but fails to evaluate tactical proficiency in realistic combat scenarios.

\paragraph{Focus on MARL Rather Than StarCraft II Mastery}
These environments were designed specifically to advance MARL algorithms rather than to develop systems that can master StarCraft II gameplay. Consequently, they prioritize properties beneficial for reinforcement learning (like simplified action spaces and reward structures) over faithful reproduction of the tactical challenges that make StarCraft II compelling.

Our AVACraft environment addresses these limitations by preserving the rich tactical depth of StarCraft II micro-management. We maintain full unit abilities, support diverse unit compositions, create realistic combat scenarios, and—most importantly—align AI perception with human gameplay experience through RGB visual inputs and natural language observations. This approach enables the development of agents that can execute sophisticated tactical maneuvers involving ability timing, positioning, and multi-unit coordination that more closely resemble human gameplay.

\newpage
\section{Hyperparameters and Reproducibility}
\label{app:hyperparameters}

\subsection{MARL Hyperparameters}

All MARL experiments use the following configuration unless otherwise noted:

\begin{table}[H]
\centering
\small
\caption{MARL training hyperparameters.}
\begin{tabular}{@{}ll@{}}
\toprule
\textbf{Parameter} & \textbf{Value} \\
\midrule
\multicolumn{2}{l}{\textit{Visual Backbone (Swin-Tiny)}} \\
Architecture & Swin-Tiny (27.5M params) \\
Input resolution (screen) & $160 \times 120$ \\
Input resolution (minimap) & $32 \times 32$ \\
Feature dimension & 768 \\
\midrule
\multicolumn{2}{l}{\textit{Text Encoder (for Vision+Text mode)}} \\
Model & GTE-Base \\
Embedding dimension & 768 \\
Fusion method & Learned linear projection \\
\midrule
\multicolumn{2}{l}{\textit{Training}} \\
Total timesteps & 5,000,000 \\
Batch size & 32 (episodes) \\
Learning rate & $5 \times 10^{-4}$ \\
Optimizer & Adam \\
Discount factor $\gamma$ & 0.99 \\
$\epsilon$-greedy (value-based) & 1.0 $\to$ 0.05 (linear, 50K steps) \\
Replay buffer size & 5,000 episodes \\
Target update interval & 200 episodes \\
\midrule
\multicolumn{2}{l}{\textit{MAPPO/IPPO Specific}} \\
Clip parameter & 0.2 \\
GAE $\lambda$ & 0.95 \\
Entropy coefficient & 0.01 \\
Number of epochs & 5 \\
Mini-batch count & 1 \\
\midrule
\multicolumn{2}{l}{\textit{Hardware}} \\
GPUs & 2$\times$ NVIDIA A100 40GB \\
Training time (3m, 5M steps) & $\sim$65--80 hours \\
\bottomrule
\end{tabular}
\end{table}

\subsection{VLM Hyperparameters}

\begin{table}[H]
\centering
\small
\caption{VLM inference hyperparameters.}
\begin{tabular}{@{}ll@{}}
\toprule
\textbf{Parameter} & \textbf{Value} \\
\midrule
\multicolumn{2}{l}{\textit{Proprietary Models}} \\
Temperature & 0.7 \\
Top-$p$ & 0.95 \\
Max tokens (per decision) & 1,024 \\
Image resolution & $1920 \times 1080$ (downscaled to API limits) \\
Decision frequency & 2 Hz \\
API retry strategy & 3 retries, exponential backoff \\
\midrule
\multicolumn{2}{l}{\textit{Open-source Models (Qwen3-VL)}} \\
Temperature & 0.7 \\
Top-$p$ & 0.95 \\
Max tokens (per decision) & 1,024 \\
Quantization & BF16 \\
Hardware & 1$\times$ NVIDIA A100 80GB \\
\midrule
\multicolumn{2}{l}{\textit{RAG Knowledge Base}} \\
Embedding model & GTE-Base \\
Retrieval top-$k$ & 3 \\
Knowledge sources & Unit stats, matchup data, pro replays \\
\bottomrule
\end{tabular}
\end{table}

\newpage
\section{Pseudocode}
\label{sec:pseudocode}
\begin{algorithm}[H]
    \caption{AVA Decision Pipeline for AVACraft}
    \label{alg:pipeline}
    \textbf{Input}: StarCraft II environment $env$, History buffer size $H$
    \begin{algorithmic}[1]
        \STATE Initialize AVACraft environment and get initial observation $o_0 = (I_0, T_0, U_0) = env.reset()$
        \STATE Initialize history buffer $\mathcal{H}$, total reward $R=0$
        \WHILE{$env$ is not terminated}
            \STATE // Stage 1: Micro-skill Planning
            \STATE Generate skill plan $S_t = \text{VLM}_{\text{plan}}(o_t, \mathcal{H})$
            
            \STATE // Stage 2: Strategic Unit Analysis
            \STATE Detect units $A_t = \text{VLM}_{\text{detect}}(I_t)$
            \FOR{each unit $u_i \in U_t$}
                \STATE Parse unit info $(id_i, type_i, pos_i, attr_i, status_i)$
            \ENDFOR
            \STATE Identify priority units $U_{\text{priority}} = \text{VLM}_{\text{analyze}}(o_t, S_t)$
            
            \STATE // Stage 3: Knowledge Integration
            \FOR{each unit $u \in U_{\text{priority}}$}
                \STATE Retrieve unit knowledge $K(u) = \text{Retrieve}(type_u)$
            \ENDFOR
            
            \STATE // Stage 4: Action Generation
            \STATE Initialize action set $a_t = \{\}$
            \FOR{each friendly unit $i$}
                \IF{$i$ should attack}
                    \STATE Add $(i,j) \in \mathcal{A}_{attack}$ to $a_t$ for target unit $j$
                \ELSIF{$i$ should move}
                    \STATE Add $(i,x,y) \in \mathcal{A}_{move}$ or $(i,d)$ to $a_t$
                \ELSIF{$i$ should use ability}
                    \STATE Add $(i,\text{ability},\text{target}) \in \mathcal{A}_{ability}$ to $a_t$
                \ENDIF
            \ENDFOR
            
            \STATE // Execute action and update
            \STATE Get the reward and next observation: $r_t, o_{t+1} = env.step(a_t)$
            \STATE Update history buffer $\mathcal{H}$
            \STATE $R \gets R + r_t$
            \STATE $o_t \gets o_{t+1}$
            
            \IF{Victory or Defeat or TimeLimit}
                \STATE break
            \ENDIF
        \ENDWHILE
        \STATE \textbf{return} total reward $R$
    \end{algorithmic}
\end{algorithm}

\newpage
\section{Map Details}
\label{map_details}
Our \textbf{AVACraft} environment features a diverse collection of 21 specialized maps, systematically categorized based on player count and ability usage capabilities. These maps originate from three primary sources: SMAC-based maps redesigned from the StarCraft Multi-Agent Challenge framework, original maps specifically designed for AVA evaluation, and selected scenarios adapted from the LLM-PySC2 framework\footnote{\url{https://github.com/NKAI-Decision-Team/LLM-PySC2}}.

Each map is meticulously designed to evaluate specific aspects of tactical proficiency and strategic decision-making:

\begin{itemize}
    \item \textbf{Unit Control}: Assessment of fundamental micromanagement capabilities
    \item \textbf{Multi-Unit Coordination}: Evaluation of strategic control over heterogeneous unit compositions
    \item \textbf{Terrain Usage}: Testing of positional awareness and environmental exploitation
    \item \textbf{Kiting}: Assessment of dynamic hit-and-run tactical execution
    \item \textbf{Split}: Evaluation of unit distribution strategies under enemy threats
    \item \textbf{Ability Usage}: Testing of ability timing optimization and target prioritization
\end{itemize}

\begin{table*}[t]
\caption{Single player maps without ability usage.}
\centering
\resizebox{\linewidth}{!}{
\begin{tabular}{@{}p{4cm}ccccccp{5cm}c@{}}
\toprule
\textbf{Map Name} & \textbf{Unit} & \textbf{Multi} & \textbf{Terrain} & \textbf{Kiting} & \textbf{Split} & \textbf{Mirror} & \multirow{2}{*}{\textbf{Units}} & \multirow{2}{*}{\textbf{Source}} \\
& \textbf{Control} & \textbf{Unit} & \textbf{Usage} & & & \textbf{Match} & & \\
\midrule
2c\_vs\_64zg & \checkmark & \checkmark & \checkmark & \checkmark & & & Player: 2 Colossi \newline Enemy: 64 Zerglings & SMAC \\
\midrule
2m\_vs\_1z & \checkmark & \checkmark & & & & & Player: 2 Marines \newline Enemy: 1 Zealot & SMAC \\
\midrule
2s\_vs\_1sc & \checkmark & \checkmark & & & & & Player: 2 Stalkers \newline Enemy: 1 Spinecrawler & SMAC \\
\midrule
3s\_vs\_3z & \checkmark & \checkmark & & & & & Player: 3 Stalkers \newline Enemy: 3 Zealots & SMAC \\
\midrule
6r\_vs\_8z & \checkmark & \checkmark & \checkmark & \checkmark & & & Player: 6 Reapers \newline Enemy: 8 Zealots & NEW \\
\midrule
8m1mv\_vs\_2st & \checkmark & \checkmark & & & & & Player: 8 Marines, 1 Medivac \newline Enemy: 2 Siege Tanks & NEW \\
\midrule
8m2st\_vs\_35zg4b & \checkmark & \checkmark & \checkmark & & & & Player: 8 Marines, 2 Siege Tanks \newline Enemy: 35 Zerglings, 4 Banelings & NEW \\
\midrule
8m\_vs\_2pc1wp & \checkmark & & & & & & Player: 8 Marines \newline Enemy: 1 Warp Prism, 2 Photon Cannons & NEW \\
\midrule
2s3z & \checkmark & \checkmark & \checkmark & & & \checkmark & Player: 2 Stalkers, 3 Zealots \newline Enemy: 2 Stalkers, 3 Zealots & SMAC \\
\midrule
3m & \checkmark & \checkmark & & & & \checkmark & Player: 3 Marines \newline Enemy: 3 Marines & SMAC \\
\midrule
mixed\_units & \checkmark & \checkmark & & & & & Player: 1 Zealot, 1 Immortal, 1 Archon, 1 Stalker, 1 Phoenix \newline Enemy: 1 Marine, 1 Marauder, 1 Reaper, 1 Hellbat, 1 Medivac, 1 Viking (Assault), 1 Ghost, 1 Banshee & NEW \\
\bottomrule
\end{tabular}}
\end{table*}

\begin{table*}[t]
\caption{Single player maps with ability usage.}
\centering
\resizebox{\linewidth}{!}{
\begin{tabular}{@{}cccccccp{5cm}c@{}}
\toprule
\textbf{Map Name} & \textbf{Unit} & \textbf{Multi} & \textbf{Terrain} & \textbf{Kiting} & \textbf{Split} & \textbf{Ability} & \multirow{2}{*}{\textbf{Units}} & \multirow{2}{*}{\textbf{Source}} \\
& \textbf{Control} & \textbf{Unit} & \textbf{Usage} & & & \textbf{Usage} & & \\
\midrule
8m3mr1mv1st\_mirror & \checkmark & \checkmark & & & \checkmark & \checkmark & Player: 8 Marines, 3 Marauders, 1 Medivac, 1 Siege Tank \newline Enemy: 8 Marines, 3 Marauders, 1 Medivac, 1 Siege Tank & NEW \\
\midrule
8s\_vs\_8m3mr1mv1st & \checkmark & & & & \checkmark & \checkmark & Player: 8 Stalkers \newline Enemy: 8 Marines, 3 Marauders, 1 Medivac, 1 Siege Tank & NEW \\
\midrule
8m3mr1mv1st\_vs\_5s2c & \checkmark & \checkmark & & & \checkmark & \checkmark & Player: 8 Marines, 3 Marauders, 1 Medivac, 1 Siege Tank \newline Enemy: 5 Stalkers, 2 Colossi & NEW \\
\midrule
pvz\_ht & \checkmark & \checkmark & & & & \checkmark & Player: 12 Stalkers, 1 Archon, 4 Sentries, 6 High Templars \newline Enemy: 64 Zerglings, 32 Banelings, 3 Ultralisks, 3 Queens & LLM-PYSC2 \\
\bottomrule
\end{tabular}}
\end{table*}

\begin{table*}[t]
\caption{Two player maps without ability usage.}
\centering
\resizebox{\linewidth}{!}{
\begin{tabular}{@{}cccccccp{5cm}c@{}}
\toprule
\textbf{Map Name} & \textbf{Unit} & \textbf{Multi} & \textbf{Terrain} & \textbf{Kiting} & \textbf{Split} & \textbf{Mirror} & \multirow{2}{*}{\textbf{Units}} & \multirow{2}{*}{\textbf{Source}} \\
& \textbf{Control} & \textbf{Unit} & \textbf{Usage} & & & \textbf{Match} & & \\
\midrule
MMM\_vs\_MMM & \checkmark & \checkmark & & \checkmark & \checkmark & \checkmark & Player 1: 8 Marines, 3 Marauders, 1 Medivac \newline Player 2: 8 Marines, 3 Marauders, 1 Medivac & SMAC \\
\midrule
mixed\_units\_pvp & \checkmark & \checkmark & & & & & Player 1: 1 Zealot, 1 Immortal, 1 Archon, 1 Stalker, 1 Phoenix \newline Player 2: 1 Marine, 1 Marauder, 1 Reaper, 1 Hellbat, 1 Medivac, 1 Viking (Assault), 1 Ghost, 1 Banshee & NEW \\
\midrule
terran\_mirror & \checkmark & \checkmark & & & & \checkmark & Player 1: 1 Marine, 1 Marauder, 1 Reaper, 1 Hellbat, 1 Medivac, 1 Viking (Assault), 1 Ghost, 1 Banshee \newline Player 2: 1 Marine, 1 Marauder, 1 Reaper, 1 Hellbat, 1 Medivac, 1 Viking (Assault), 1 Ghost, 1 Banshee & NEW \\
\bottomrule
\end{tabular}}
\end{table*}

\begin{table*}[t]
\caption{Two player maps with ability usage.}
\centering
\resizebox{\linewidth}{!}{
\begin{tabular}{@{}cccccccp{5cm}c@{}}
\toprule
\textbf{Map Name} & \textbf{Unit} & \textbf{Multi} & \textbf{Terrain} & \textbf{Kiting} & \textbf{Split} & \textbf{Ability} & \multirow{2}{*}{\textbf{Units}} & \multirow{2}{*}{\textbf{Source}} \\
& \textbf{Control} & \textbf{Unit} & \textbf{Usage} & & & \textbf{Usage} & & \\
\midrule
7s\_vs\_11m1mv1st & \checkmark & & & \checkmark & \checkmark & \checkmark & Player 1: 7 Stalkers \newline Player 2: 11 Marines, 1 Medivac, 1 Siege Tank & NEW \\
\midrule
8s\_vs\_8m3mr1mv1st\_pvp & \checkmark & & & \checkmark & \checkmark & \checkmark & Player 1: 8 Stalkers \newline Player 2: 8 Marines, 3 Marauders, 1 Medivac, 1 Siege Tank & NEW \\
\midrule
8m3mr1mv1st\_mirror\_pvp & \checkmark & \checkmark & & \checkmark & \checkmark & \checkmark & Player 1: 8 Marines, 3 Marauders, 1 Medivac, 1 Siege Tank \newline Player 2: 8 Marines, 3 Marauders, 1 Medivac, 1 Siege Tank & NEW \\
\bottomrule
\end{tabular}}
\end{table*}

\section{Evaluation Metrics}
\label{app:evaluation_metrics}

We define the three metrics used in the human evaluation of AVA and MARL agents, each rated on a 1--5 scale:

\begin{itemize}
    \item \textbf{Game Bug Exploitation}: Measures whether the agent exploits game bugs, particularly vulnerabilities in SMAC's built-in AI, which uses a flawed strategy of attacking only the enemy's spawn point and stopping if the enemy moves out of range or beyond attack distance (1 = frequent exploitation, 5 = no exploitation).
    \item \textbf{Reasoning Coherence}: Evaluates whether the agent's decisions are logical, incorporating StarCraft II game knowledge (e.g., unit matchups) and operational skills (e.g., positioning, targeting) (1 = illogical, 5 = perfect logic).
    \item \textbf{Human Similarity}: Assesses how closely the agent's strategies resemble human play, including techniques like hit-and-run tactics and multi-unit coordination (e.g., combined-arms strategies) (1 = unlike human, 5 = completely human-like).
\end{itemize}

Evaluations were conducted in a blinded setting: participants watched recorded replays without being informed whether the controlling agent was MARL-based or VLM-based. Each participant evaluated 10 replays (5 per paradigm) in randomized order. Statistical significance was assessed using the Mann-Whitney U test due to the ordinal nature of Likert scale data and the small sample size.

\section{Open-Source VLM Extended Results}
\label{app:opensource_vlm}

To ensure reproducibility without proprietary API dependencies, we provide extended results for open-source Qwen3-VL models across all evaluated scenarios.

\begin{table}[H]
\centering
\small
\caption{Extended open-source VLM results. Win rates (\%) over 20 episodes.}
\begin{tabular}{@{}lcc@{}}
\toprule
\textbf{Scenario} & \textbf{Qwen3-VL-8B} & \textbf{Qwen3-VL-30B} \\
\midrule
vlm\_attention\_1 & 40 & \textbf{90} \\
3m & 40 & 50 \\
2m\_vs\_1z & 5 & 15 \\
mixed\_units & 35 & 60 \\
2s3z & 10 & 20 \\
3s\_vs\_3z & 5 & 15 \\
2s\_vs\_1sc & 0 & 0 \\
pvz\_ht & 10 & 20 \\
8m2st\_vs\_35zg4b & 15 & 30 \\
8m1mv\_vs\_2st & 0 & 5 \\
8m\_vs\_2pc1wp & 0 & 5 \\
6r\_vs\_8z & 0 & 0 \\
2c\_vs\_64zg & 0 & 0 \\
\midrule
\textbf{Average} & 12.3 & 23.8 \\
\bottomrule
\end{tabular}
\end{table}

The clear performance jump from 8B (12.3\% average) to 30B (23.8\% average) demonstrates AVACraft's ability to measure VLM scaling laws in spatial reasoning tasks. Notably, Qwen3-VL-30B achieves 90\% on vlm\_attention\_1 (a focused targeting scenario), confirming that the environment mechanics and API pipelines are entirely functional and that the 0\% win rates on complex maps reflect genuine capability ceilings of current VLMs.

\section{Case of Study}

\begin{figure}[t]
    \centering
    \includegraphics[width=\linewidth]{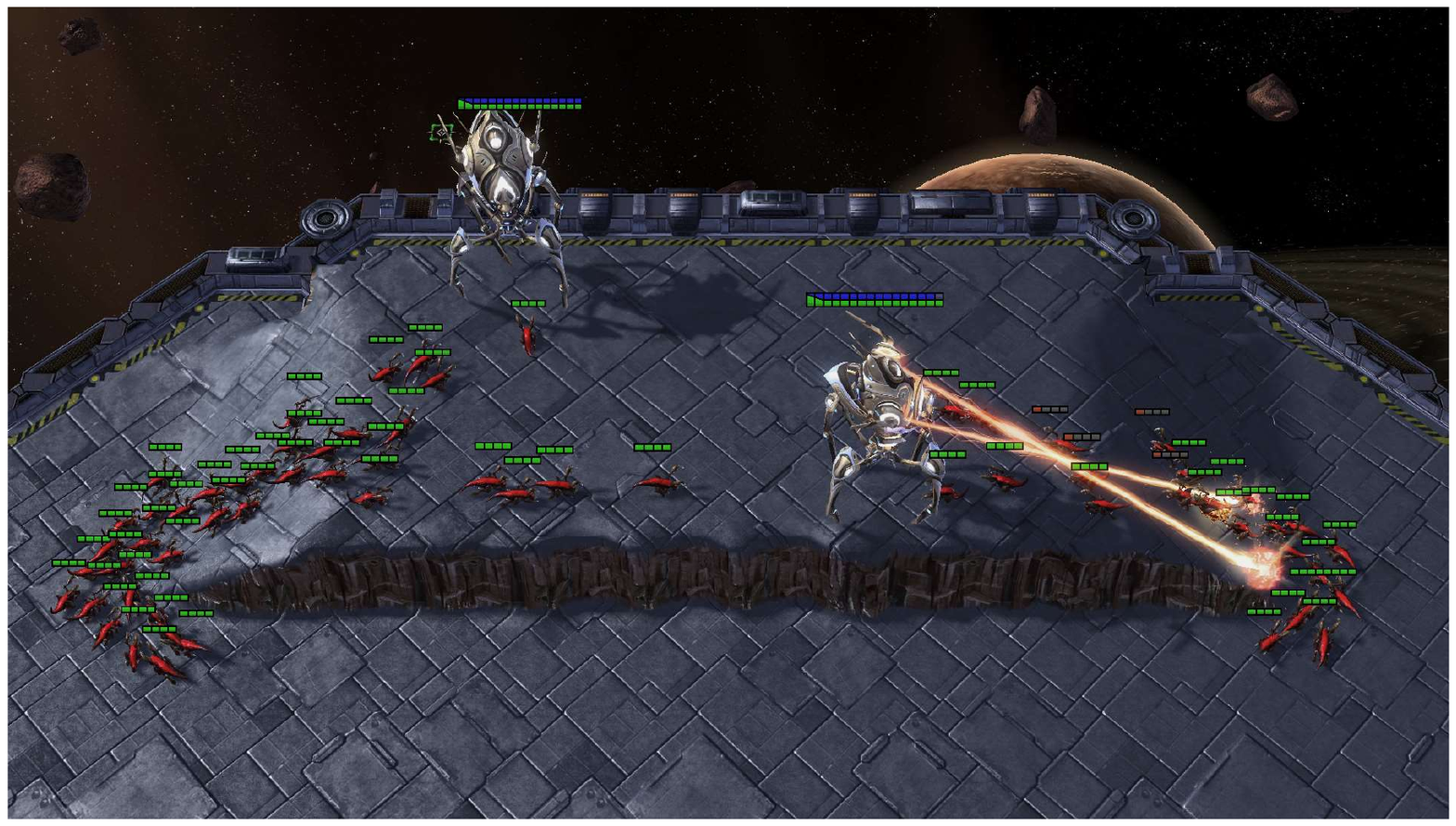}
    \caption{Original RGB observation of battlefield situation in the Colossi vs Zerglings scenario.}
    \label{fig:colossus_aoe_01}
\end{figure}

\begin{figure}[t]
    \centering
    \includegraphics[width=\linewidth]{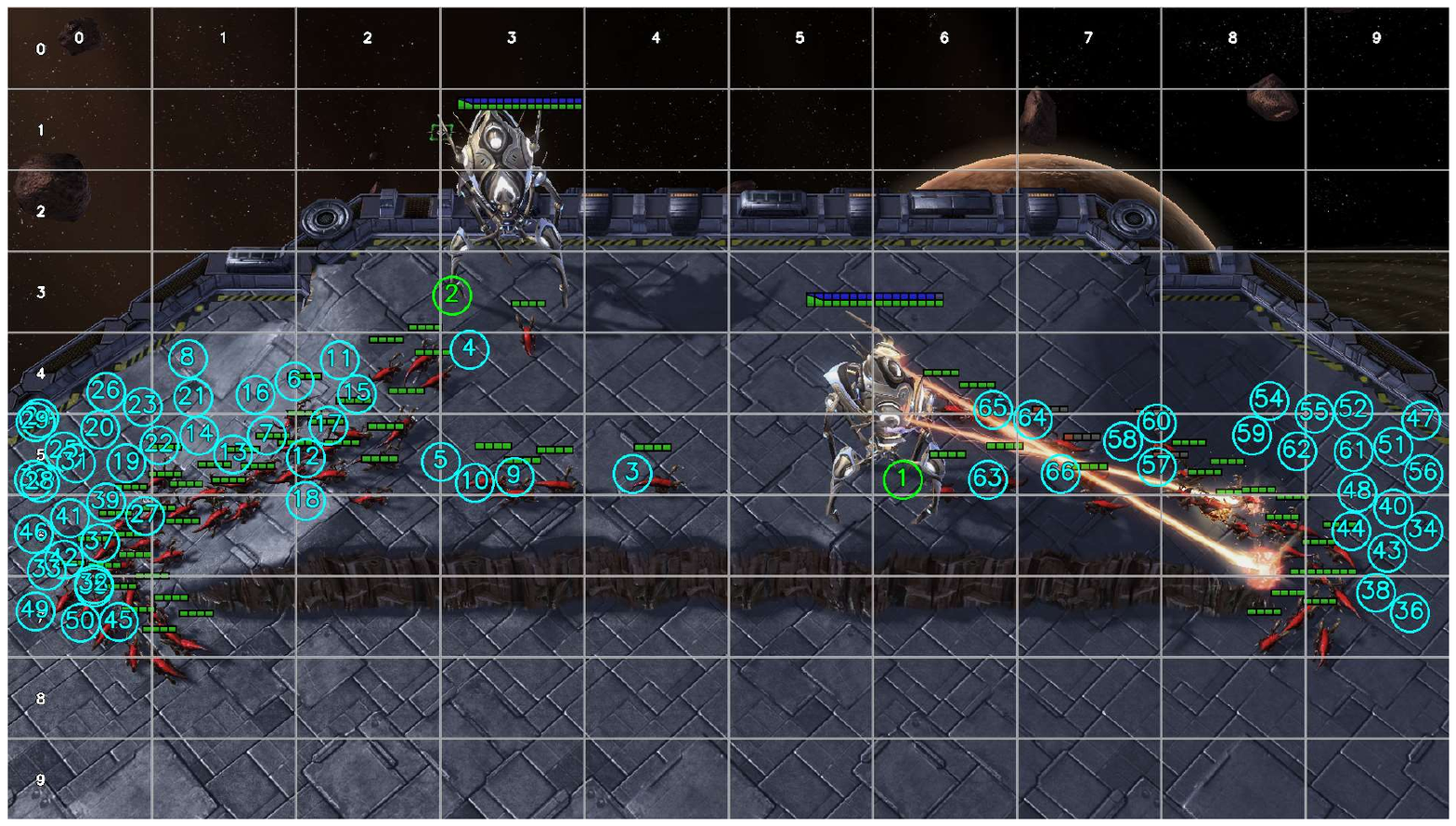}
    \caption{Annotated unit positions with unit IDs and health status.}
    \label{fig:colossus_aoe_02}
\end{figure}

Figures \ref{fig:colossus_aoe_01} and \ref{fig:colossus_aoe_02} illustrate the initial stages of AVA's decision-making process. The system begins by processing the raw RGB battlefield observation, then identifies and annotates individual units with their respective IDs and health status. This visual processing stage forms the foundation for subsequent tactical analysis.

\begin{figure}[t]
    \centering
    \includegraphics[width=\linewidth]{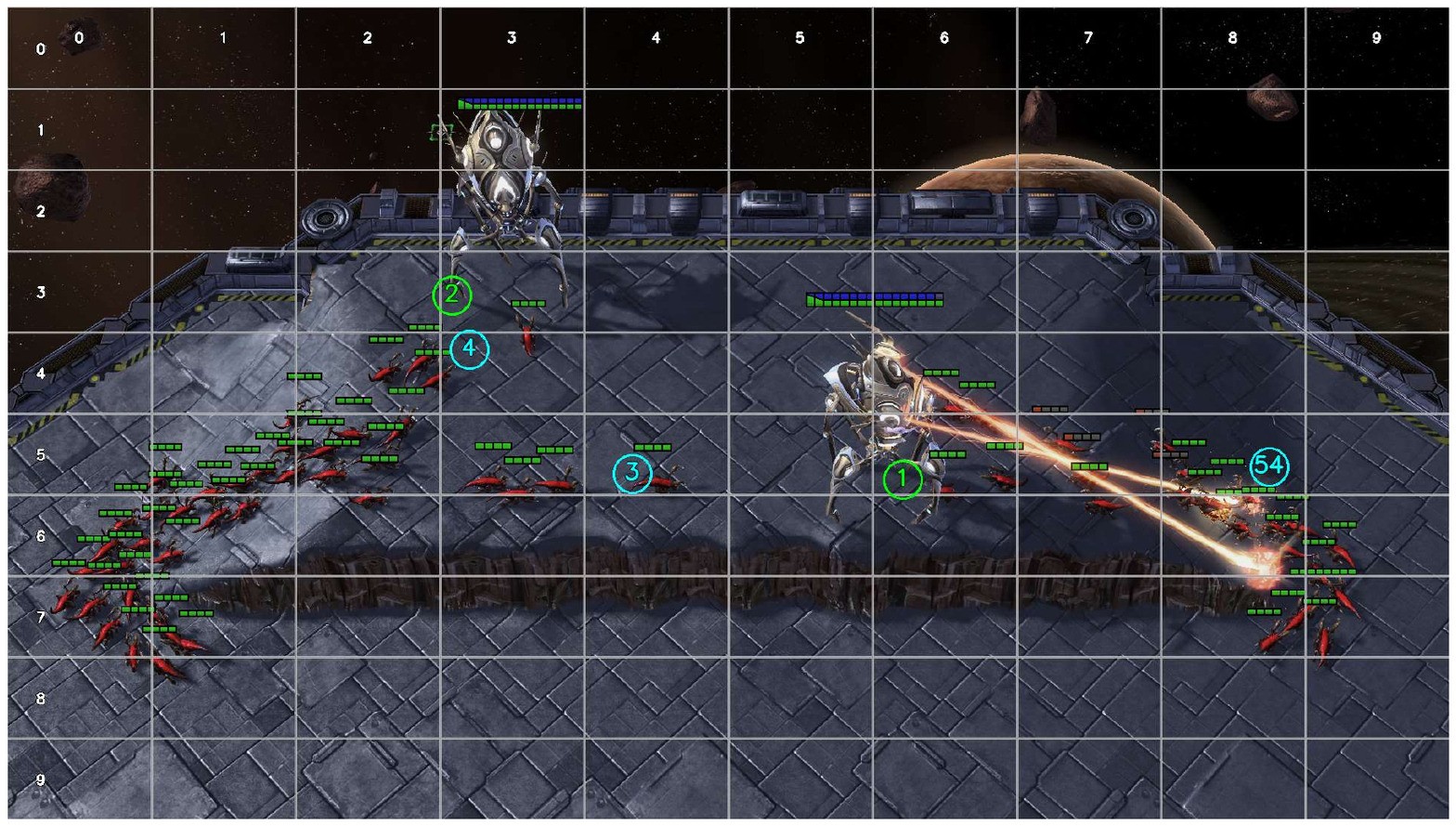}
    \caption{AVA's strategic analysis highlighting prioritized targets and optimal attack vectors.}
    \label{fig:colossus_aoe_03}
\end{figure}

Figure \ref{fig:colossus_aoe_03} demonstrates AVA's strategic decision-making capabilities. In this complex micro-management scenario, AVA identified Zergling\_52 (Tag: 54) as a priority target due to its strategic position at [2,1], where attacking it would maximize area-of-effect damage to nearby clustered units. This decision demonstrates the system's ability to not only identify low-health targets (5/35 HP) but also recognize opportunities for efficient damage distribution through Colossi's line damage mechanic. Supporting this decision, the system also identified Zergling\_1 (Tag: 3) and Zergling\_2 (Tag: 4) as secondary priority targets due to their threatening positions at [1,1] and [0,1] respectively, enabling a comprehensive control strategy that combines focus fire with positional advantage.

\begin{figure}[t]
    \centering
    \setlength{\tabcolsep}{1pt}
    \begin{tabular}{c}
        \subfloat[Initial state showing Marine/Tank positions]{
            \includegraphics[width=\linewidth]{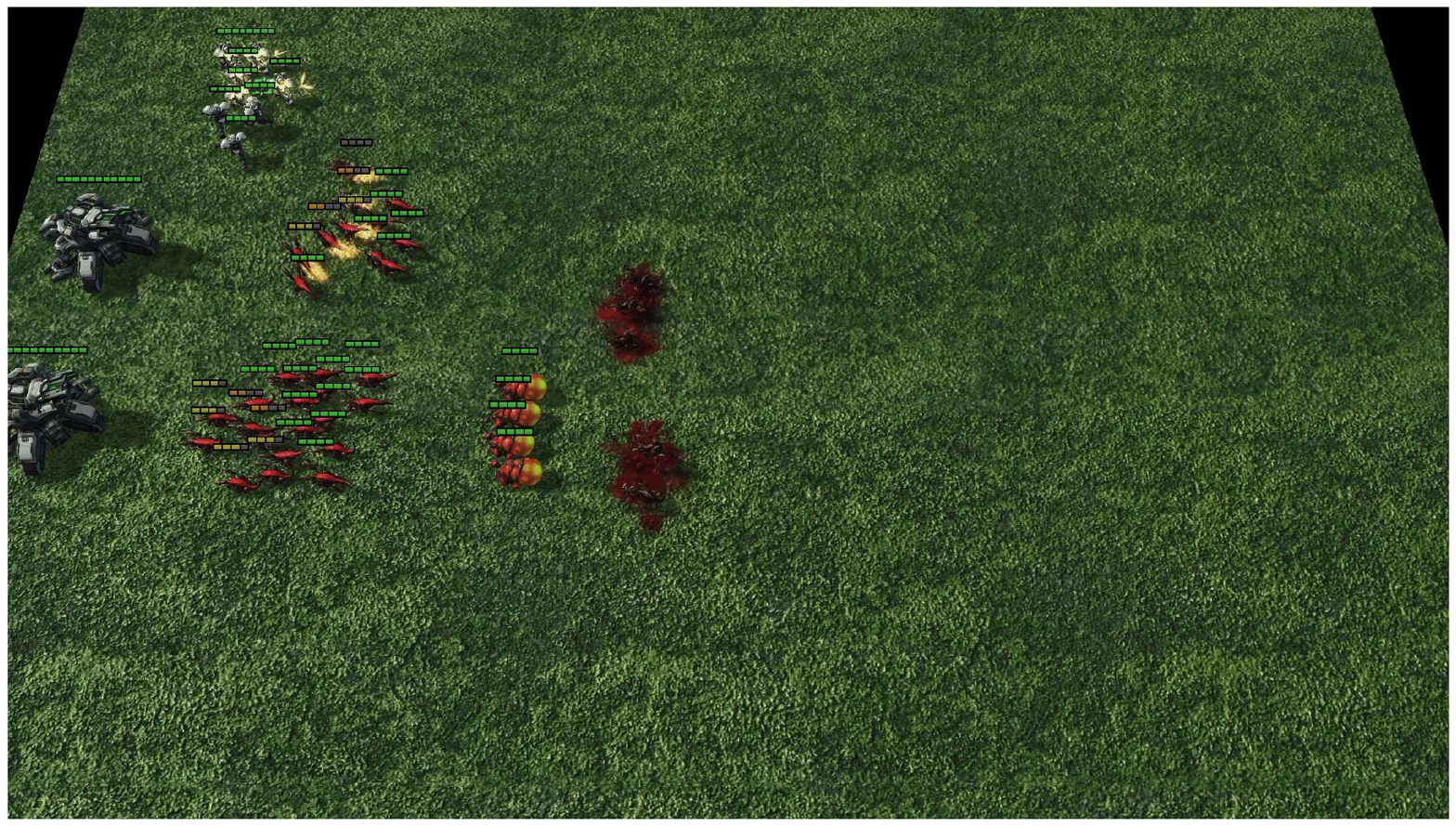}} \\
        \subfloat[VLM unit identification]{
            \includegraphics[width=\linewidth]{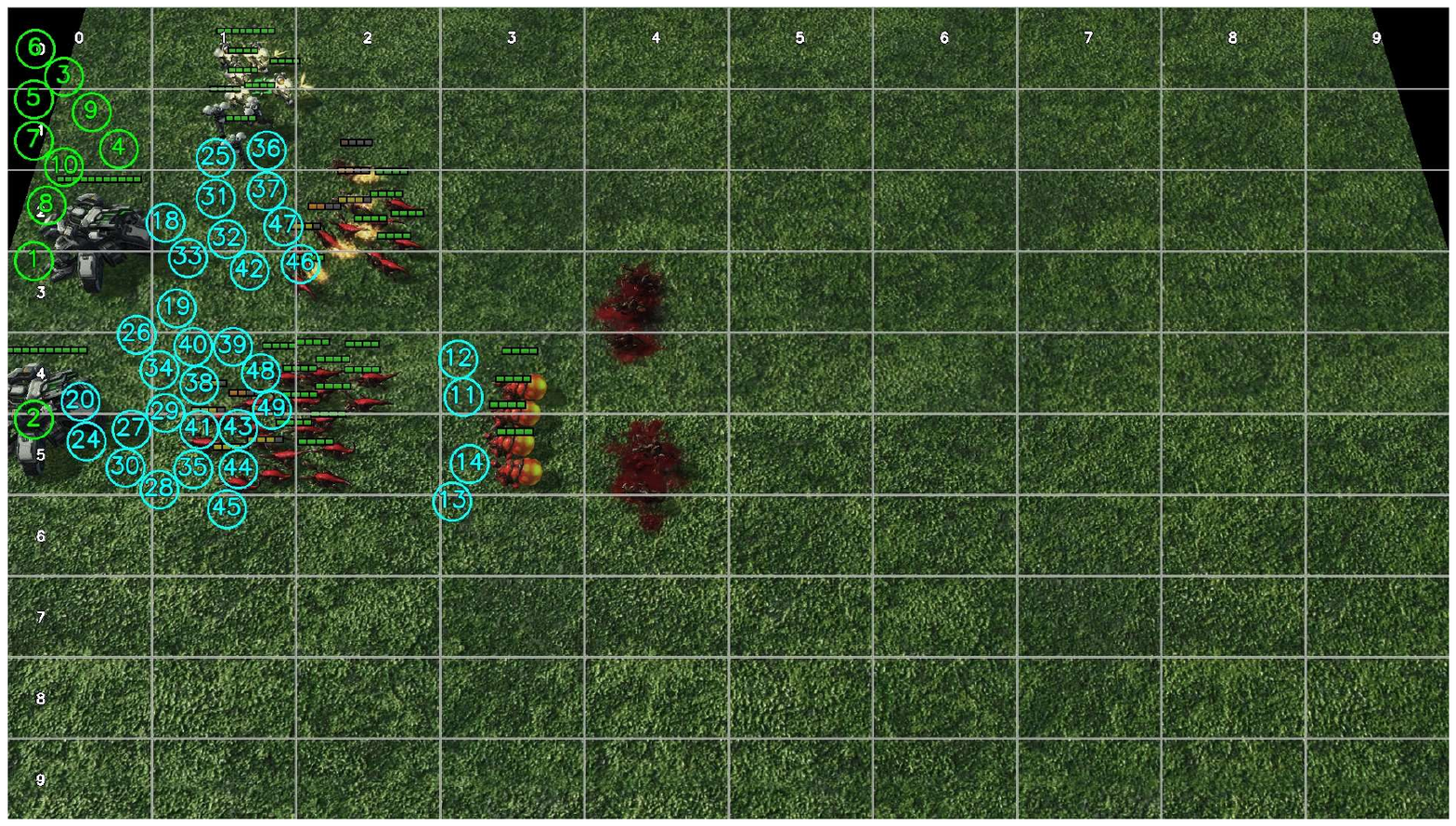}} \\
        \subfloat[Priority targeting analysis]{
            \includegraphics[width=\linewidth]{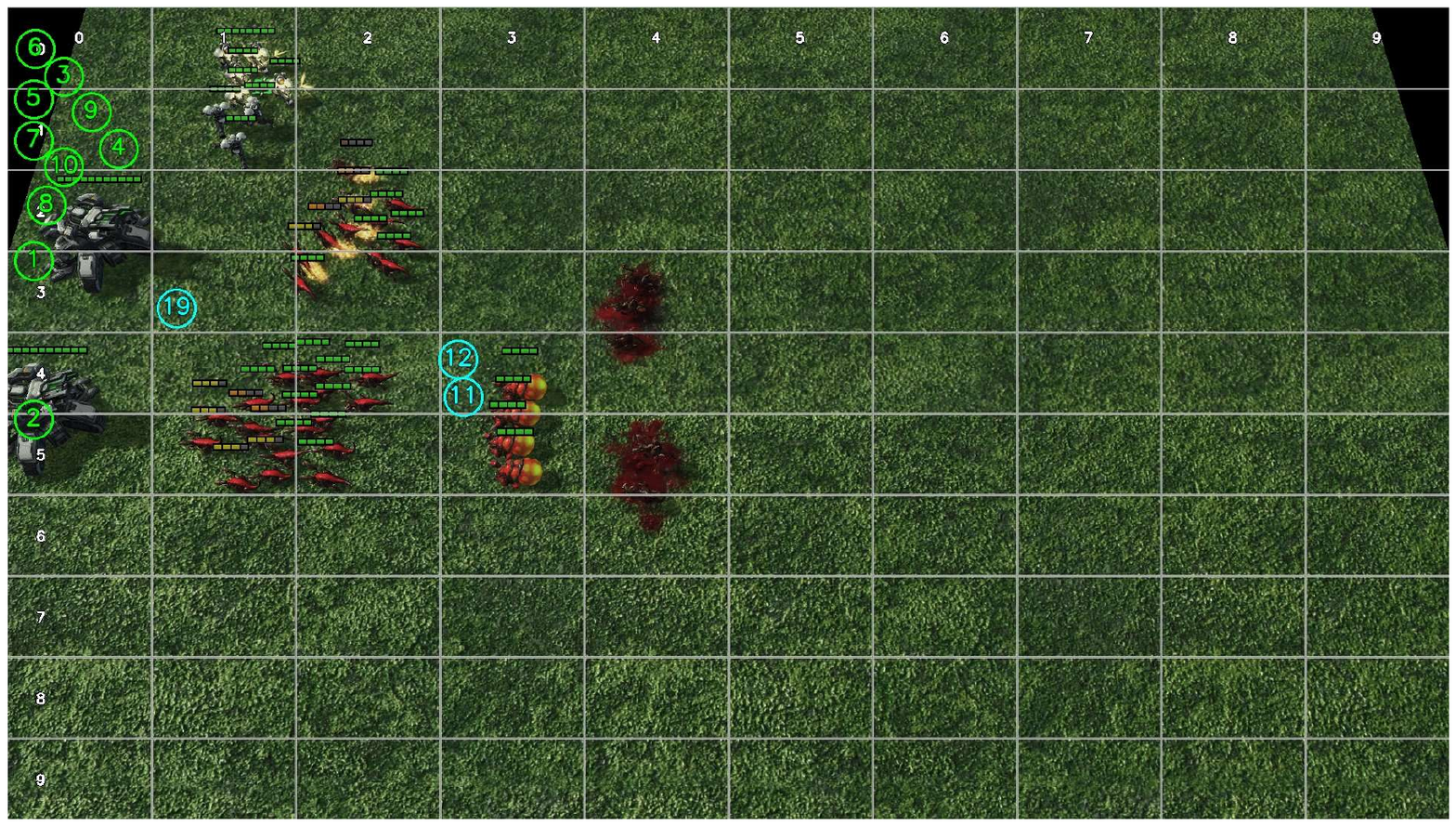}}
    \end{tabular}
    \caption{Stage 1: AVA's battlefield analysis and threat assessment in Marine/Tank vs Baneling/Zergling engagement.}
    \label{fig:combined_stage1}
\end{figure}

\begin{figure}[h]
    \centering
    \setlength{\tabcolsep}{1pt}
    \begin{tabular}{c}
        \subfloat[Marine formation adjustment]{
            \includegraphics[width=0.5\textwidth]{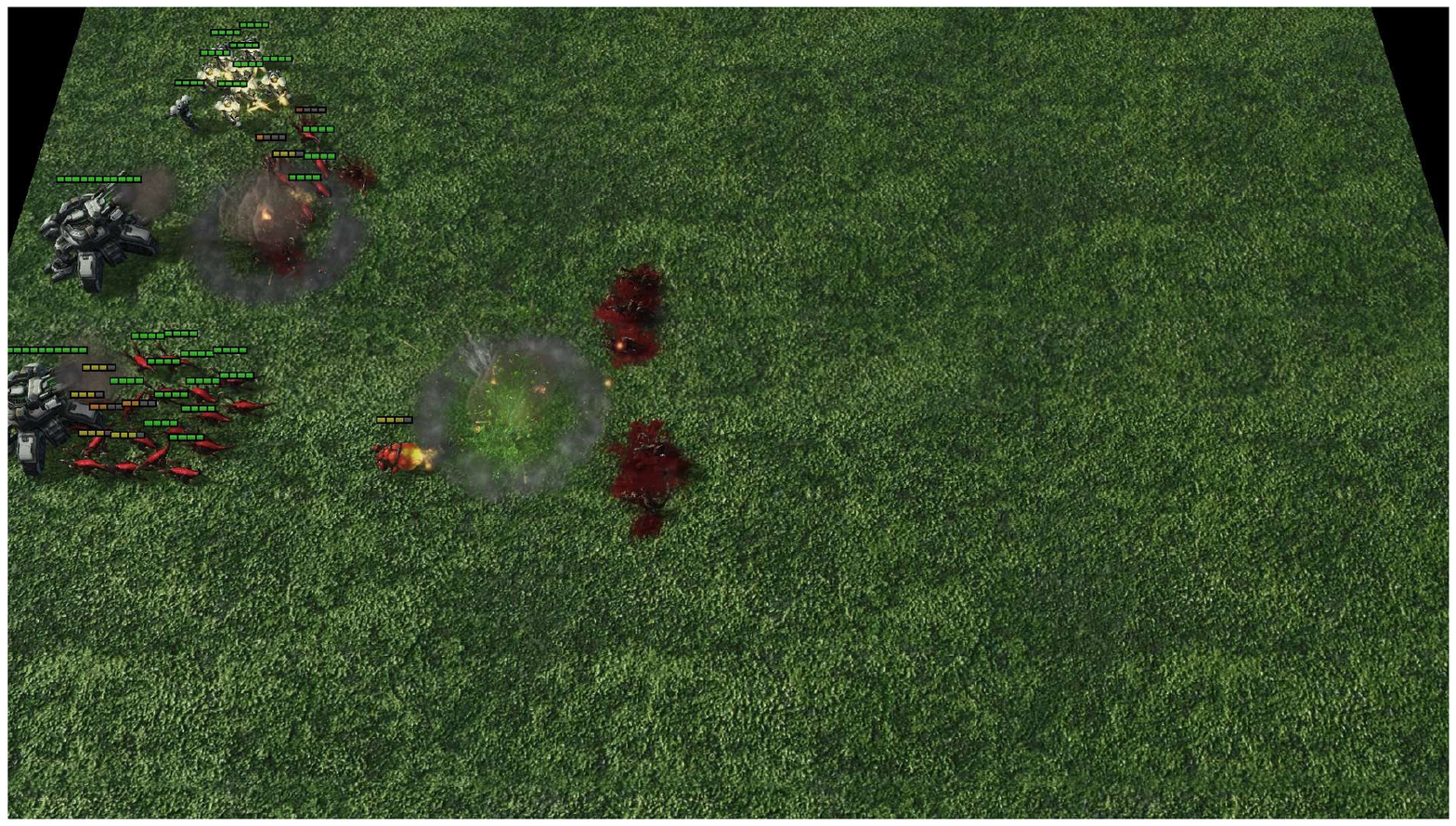}} \\
        \subfloat[Coordinated focus fire execution]{
            \includegraphics[width=0.5\textwidth]{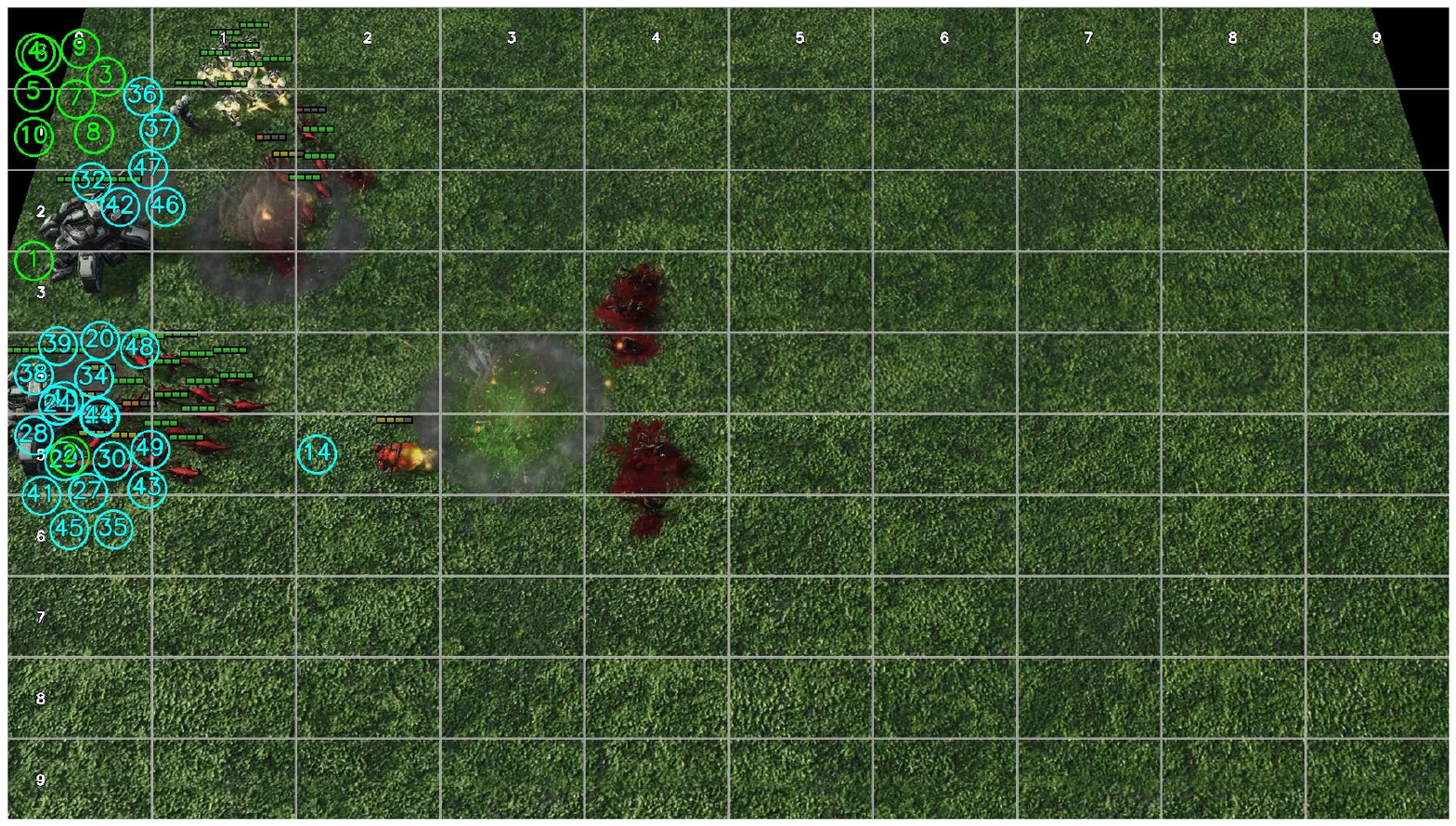}} \\
        \subfloat[Optimized Marine positioning]{
            \includegraphics[width=0.5\textwidth]{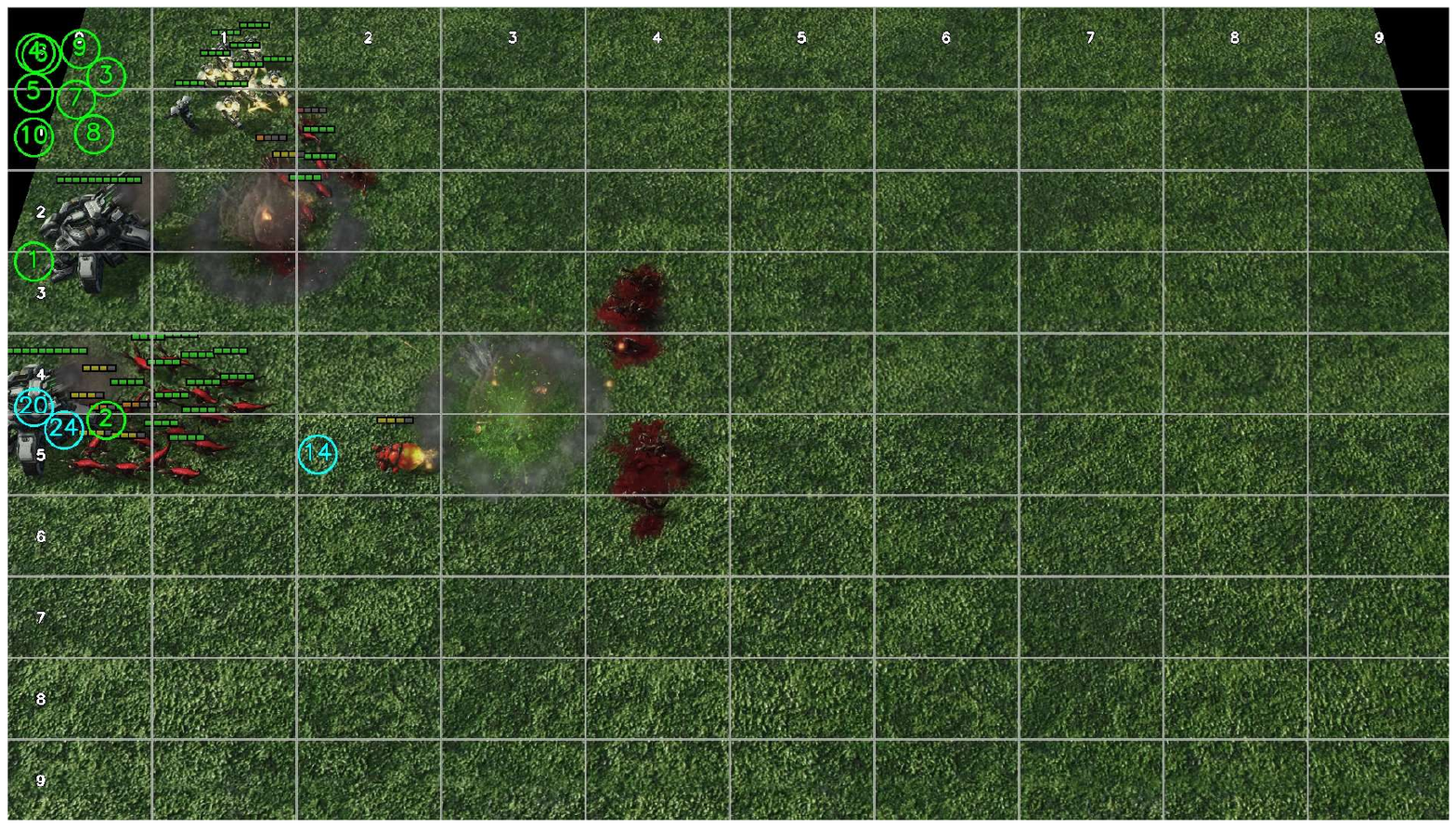}}
    \end{tabular}
    \caption{Stage 2: Tactical positioning and focus fire coordination on priority targets.}
    \label{fig:combined_stage2}
\end{figure}

\begin{figure}[h]
    \centering
    \setlength{\tabcolsep}{1pt}
    \begin{tabular}{c}
        \subfloat[Secondary target engagement]{
            \includegraphics[width=0.5\textwidth]{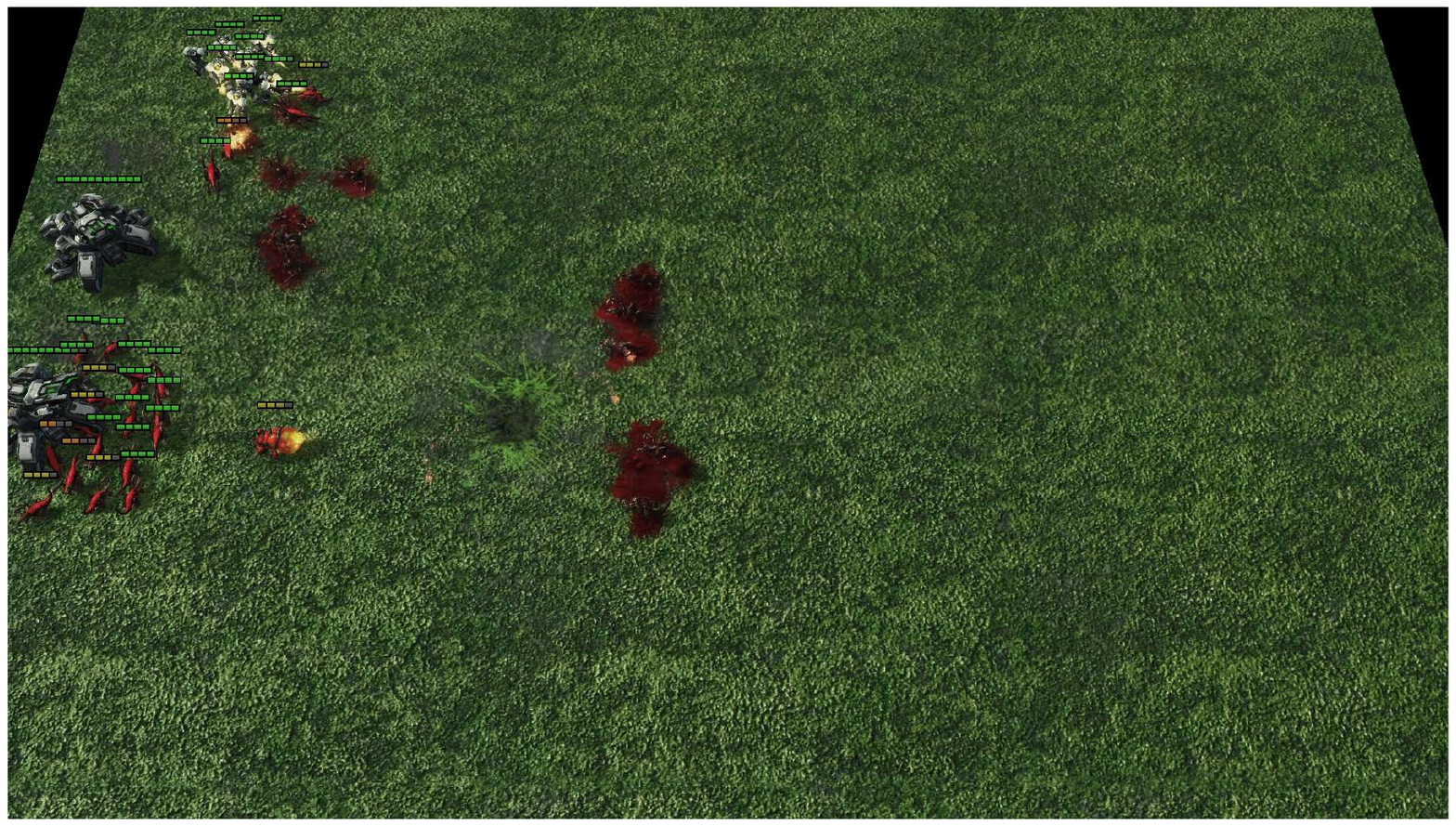}} \\
        \subfloat[Maintained spread formation]{
            \includegraphics[width=0.5\textwidth]{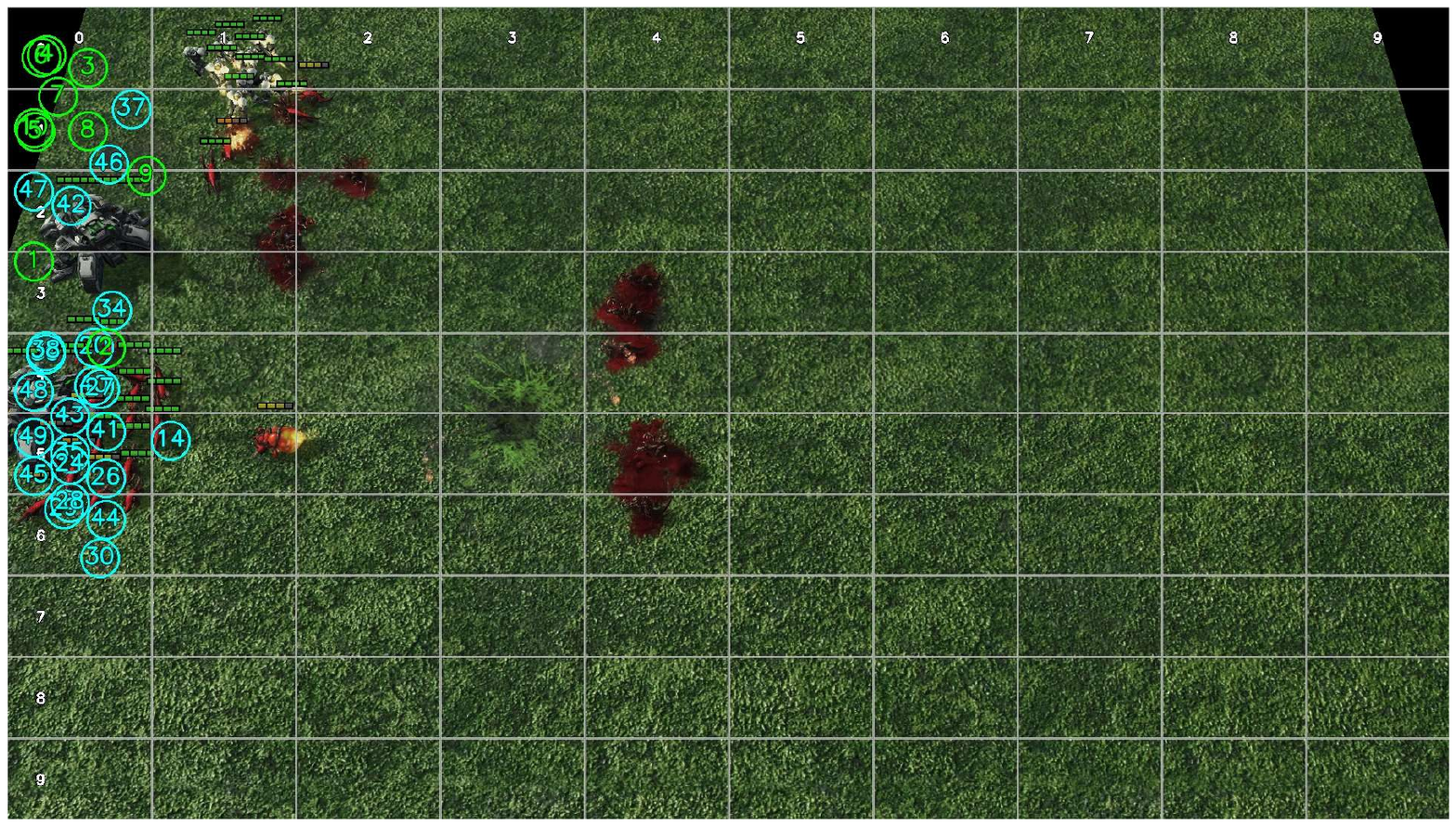}} \\
        \subfloat[Final engagement phase]{
            \includegraphics[width=0.5\textwidth]{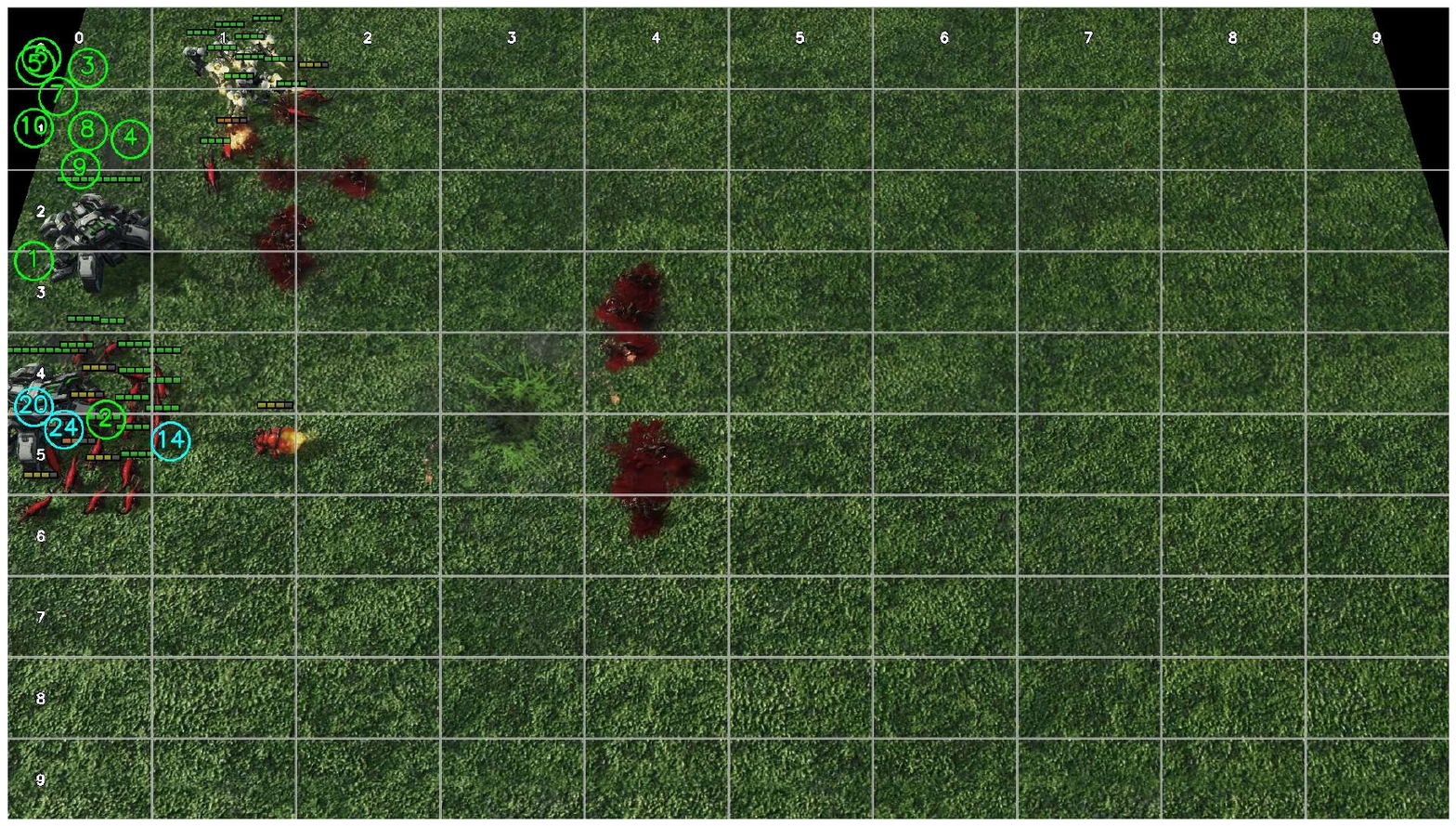}}
    \end{tabular}
    \caption{Stage 3: Sequential target elimination while maintaining strategic formation.}
    \label{fig:combined_stage3}
\end{figure}

The tactical execution depicted in Figures~\ref{fig:combined_stage1}, \ref{fig:combined_stage2}, and \ref{fig:combined_stage3} showcases AVA's sophisticated decision-making processes that emerge without explicit training. The system first performs battlefield analysis, identifying Banelings as primary threats due to their splash damage potential against clustered units. It then implements a coordinated response by strategically positioning Marines at safe distances while maintaining focus fire capabilities. Throughout the engagement, AVA demonstrates multiple micro-skills simultaneously: prioritized target selection, formation control, and adaptive positioning. This behavior closely resembles human expert gameplay strategies, highlighting AVA's ability to leverage VLM reasoning for complex tactical decision-making that would typically require extensive reinforcement or imitation learning in traditional approaches.

\begin{figure}[t]
    \centering
    \setlength{\tabcolsep}{1pt}
    \begin{tabular}{c}
        \subfloat[Initial battlefield analysis with unit annotations]{
            \includegraphics[width=\linewidth]{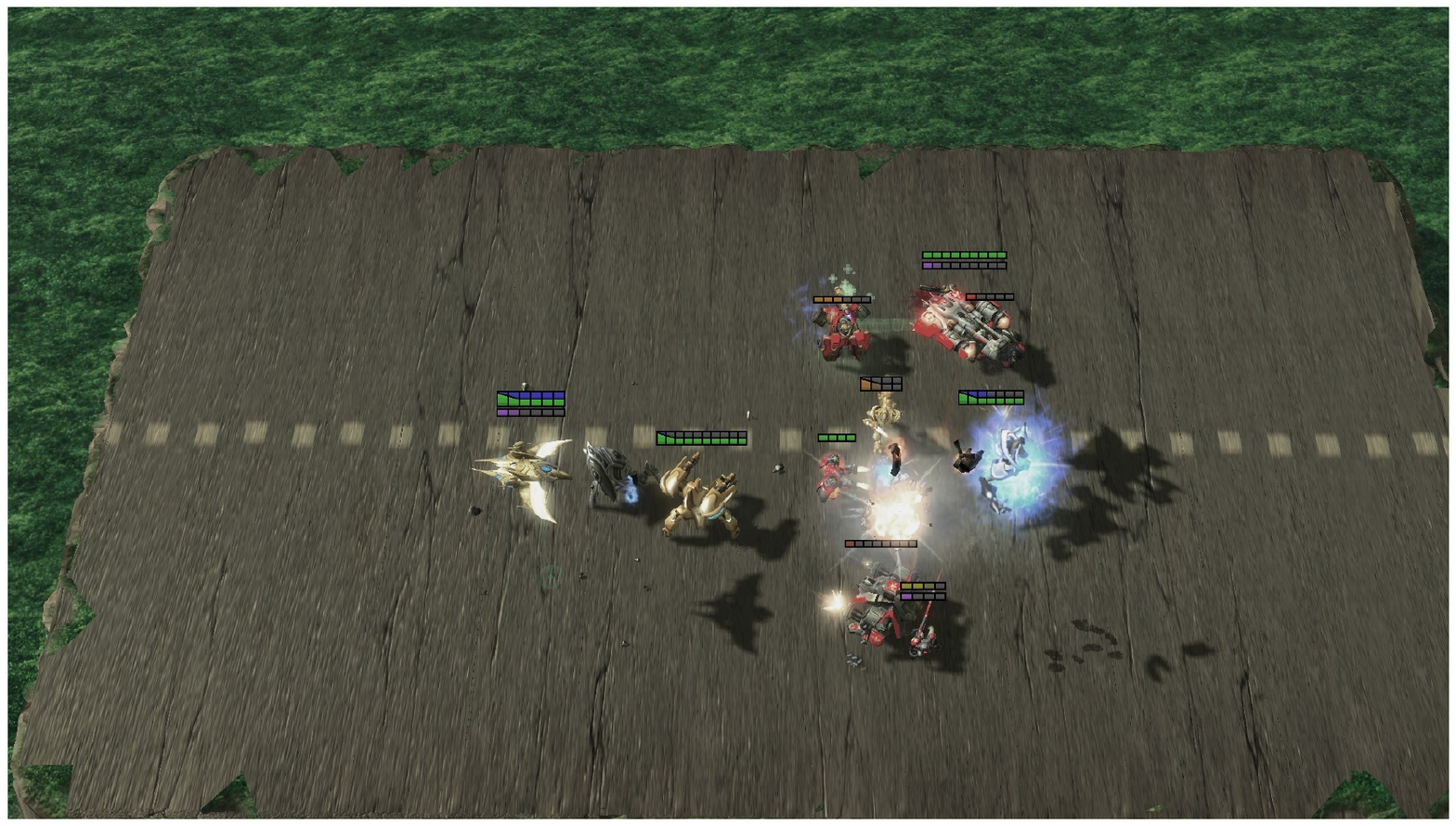}
        } \\
        \subfloat[Coordinated attack execution and positioning]{
            \includegraphics[width=\linewidth]{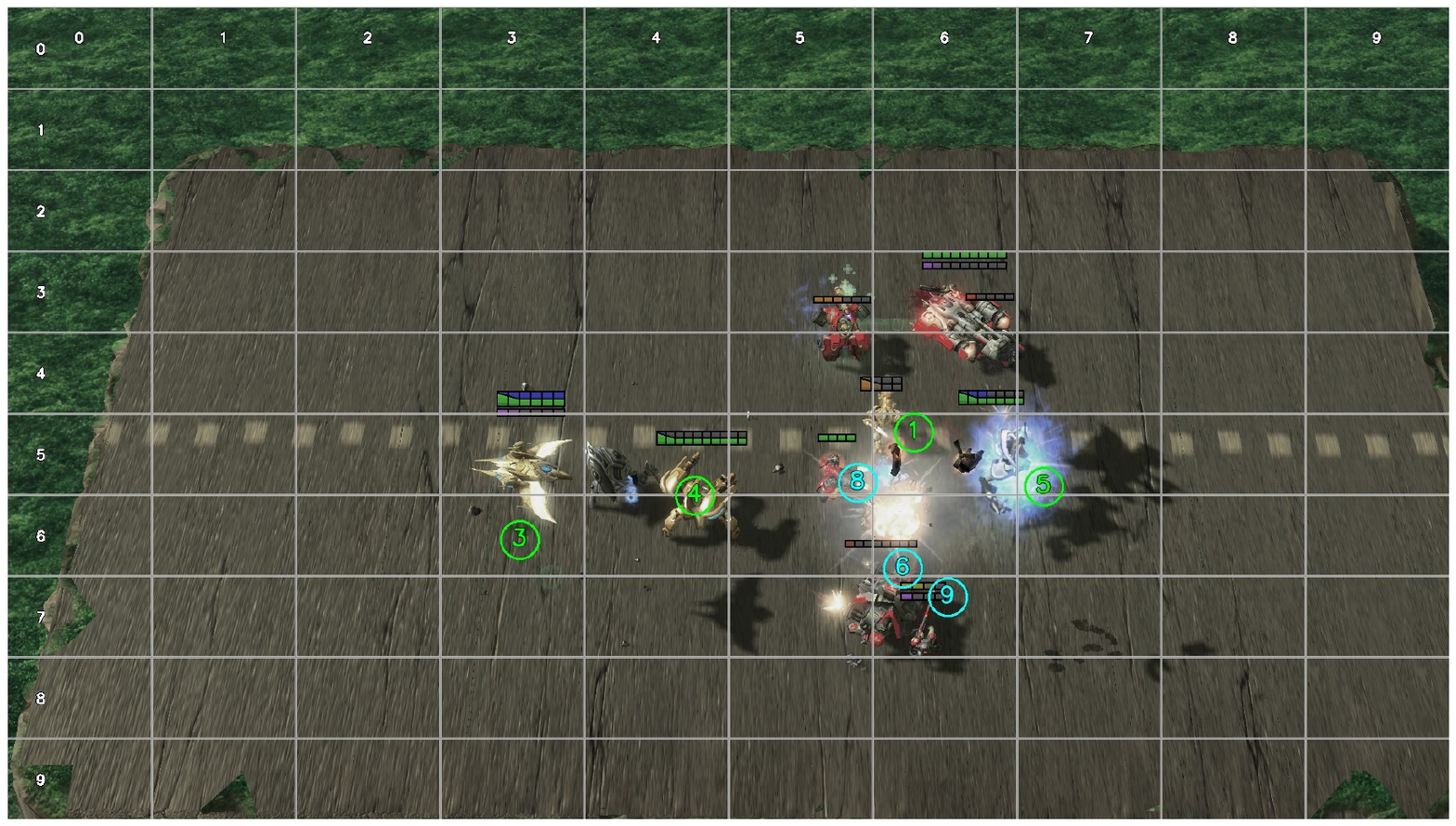}
        }
    \end{tabular}
    \caption{Multi-type unit coordination in Protoss vs Terran engagement, showing AVA's strategic targeting based on unit attributes and tactical synergies.}
    \label{fig:multi_type_cooperation}
\end{figure}

Figure \ref{fig:multi_type_cooperation} illustrates AVA's ability to coordinate heterogeneous unit compositions. In the initial analysis phase (a), the system identifies critical targets including a low-health Viking Assault (11/125 HP), an energy-rich Ghost (56 energy), and support units like Medivac. Based on this assessment, it executes a coordinated attack plan (b) where each unit is assigned optimal targets: Zealot engages the weakened Viking, Phoenix provides air superiority against Medivac, Immortal focuses on armored targets, while the Archon maintains a strategic position for battlefield control. This demonstrates VLM's understanding of unit-specific attributes (health states, energy levels, armor types) and tactical synergies in mixed-unit scenarios without requiring explicit training.

\begin{figure}[t]
    \centering
    \includegraphics[width=\linewidth]{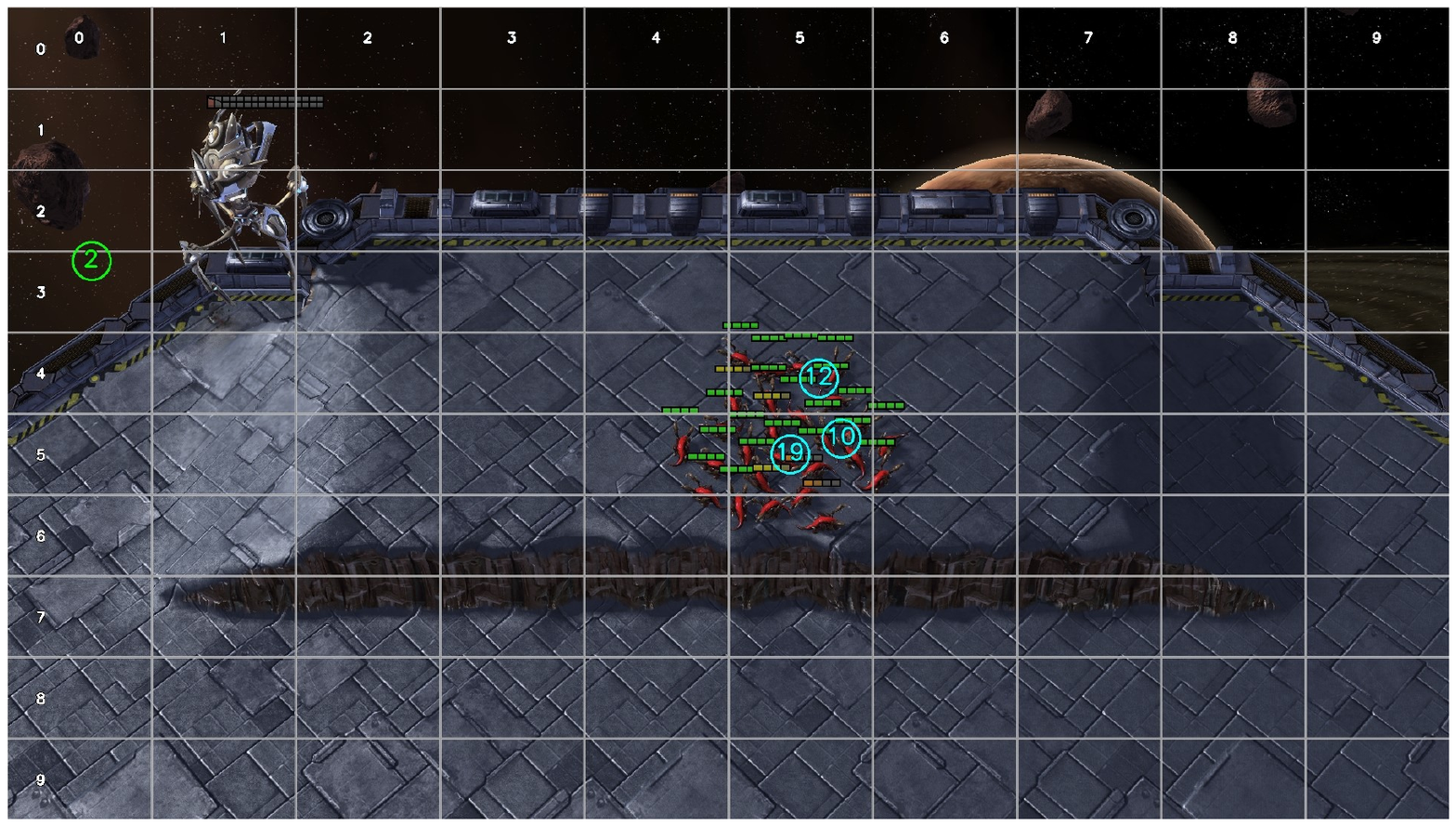}
    \caption{Tactical terrain exploitation: Colossi positioned in corner location to maximize attack range while minimizing exposure to enemy units.}
    \label{fig:colossus_in_corner}
\end{figure}

AVA demonstrates robust performance in scenarios requiring strategic target selection and basic coordination but encounters challenges with complex micro-management tasks requiring precise ability timing (as in \texttt{2s\_vs\_1sc\_vlm\_priority}) or sophisticated terrain exploitation (as in \texttt{2c\_vs\_64zg\_vlm\_priority}, Figure~\ref{fig:colossus_in_corner}). Through systematic analysis, we identified three primary limitations: (1) inconsistent spatial understanding in dense unit formations and (2) challenges in maintaining temporal consistency during high-frequency decision cycles.

\end{document}